\documentclass[a4paper]{article}

\usepackage[top=3cm, bottom=3cm, left=3cm, right=3cm, includefoot]{geometry}

\usepackage{times}
\usepackage{graphicx}
\usepackage{subfigure} 
\usepackage{natbib}
\usepackage{algorithm}
\usepackage{algorithmic}
\usepackage{hyperref}

\usepackage[cmex10]{amsmath}
\usepackage{amssymb}
\usepackage{amsfonts}
\usepackage{url}
\usepackage{bm}
\usepackage{multirow}

\usepackage{pgfplots}
\usetikzlibrary{patterns}

\renewcommand{\algorithmicrequire}{\textbf{Input:}}
\renewcommand{\algorithmicensure}{\textbf{Output:}}

\setcitestyle{authoryear,round,citesep={;},aysep={,},yysep={;}}

\renewcommand{\cite}[1]{\citep{#1}}

\newtheorem{theorem}{Theorem}

\newtheorem{lemma}{Lemma}

\newtheorem{problem}{Problem}

\renewcommand{\matrix}[1]{\ensuremath{\begin{bmatrix} #1 \end{bmatrix}}}

\newcommand{\argminl}{\mathop{\mathrm{argmin}}\nolimits}

\newcommand{\argmaxl}{\mathop{\mathrm{argmax}}\nolimits}
\newcommand{\R}{\ensuremath{\mathbb{R}}}

\makeatletter
\newcommand{\figcaption}[1]{\def\@captype{figure}\caption{#1}}
\newcommand{\tblcaption}[1]{\def\@captype{table}\caption{#1}}
\makeatother

\title{Making Tree Ensembles Interpretable:\\ A Bayesian Model Selection Approach}
\author{
  Satoshi Hara\\
    National Institute of Informatics, Japan\\
    JST, ERATO, Kawarabayashi Large Graph Project\\
   \texttt{satohara@nii.ac.jp}\\
  \\
  Kohei Hayashi\\
   National Institute of Advanced Industrial Science and Technology, Japan\\
   \texttt{hayashi.kohei@gmail.com}
}

\begin{document} 

\maketitle

\begin{abstract} 
Tree ensembles, such as random forests and boosted trees, are renowned for their high prediction performance.
However, their interpretability is critically limited due to the enormous complexity.
In this study, we present a method to make a complex tree ensemble interpretable by simplifying the model.
Specifically, we formalize the simplification of tree ensembles as a model selection problem.
Given a complex tree ensemble, we aim at obtaining the simplest representation that is essentially equivalent to the original one.
To this end, we derive a Bayesian model selection algorithm that optimizes the simplified model while maintaining the prediction performance.
Our numerical experiments on several datasets showed that complicated tree ensembles were reasonably approximated as interpretable.
\end{abstract} 

\section{Introduction}
\label{sec:intro}

Tree ensembles such as random forests~\cite{breiman2001random} and boosted trees~\cite{friedman2001greedy} are popular machine learning models, particularly for prediction tasks.
A tree ensemble builds numerous decision trees that divide an input space into a ton of tiny \emph{regions}, places their own outputs for all the regions, and makes predictions by averaging all the outputs.
Owing to the power of model averaging, their prediction performance is considerably high, and they are one of the must-try methods when dealing with real problems.
Indeed, it is reported that XGBoost~\cite{chen2016xgboost}, the state-of-the-art tree ensemble method, is one of the most popular methods in Kaggle competitions~\cite{Kaggle2016}.

\begin{figure*}[t]
	\centering
	\subfigure[Original Data]{
		\begin{minipage}[c]{0.31\textwidth}
		\includegraphics[width=0.95\textwidth]{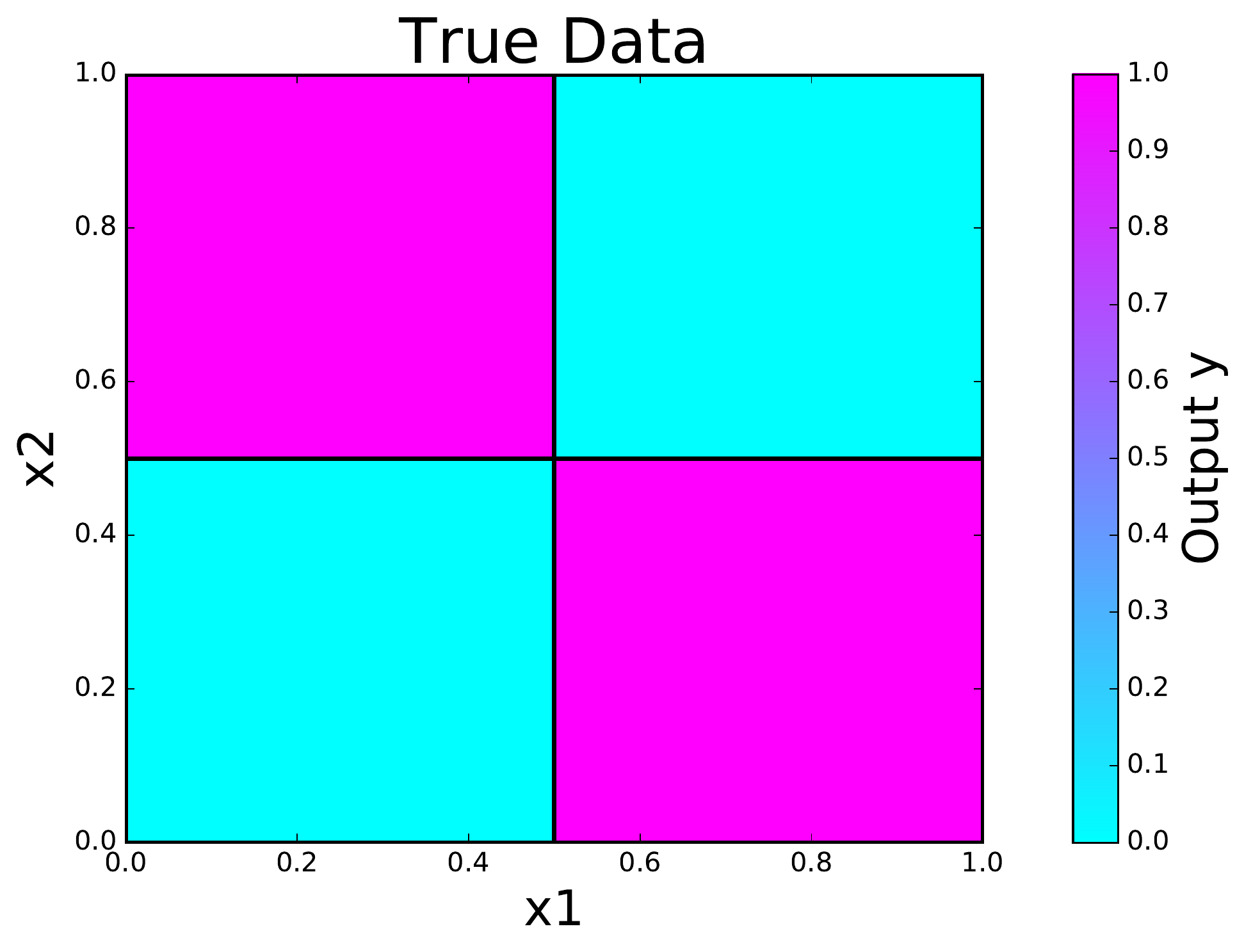}\\
		\includegraphics[width=0.95\textwidth]{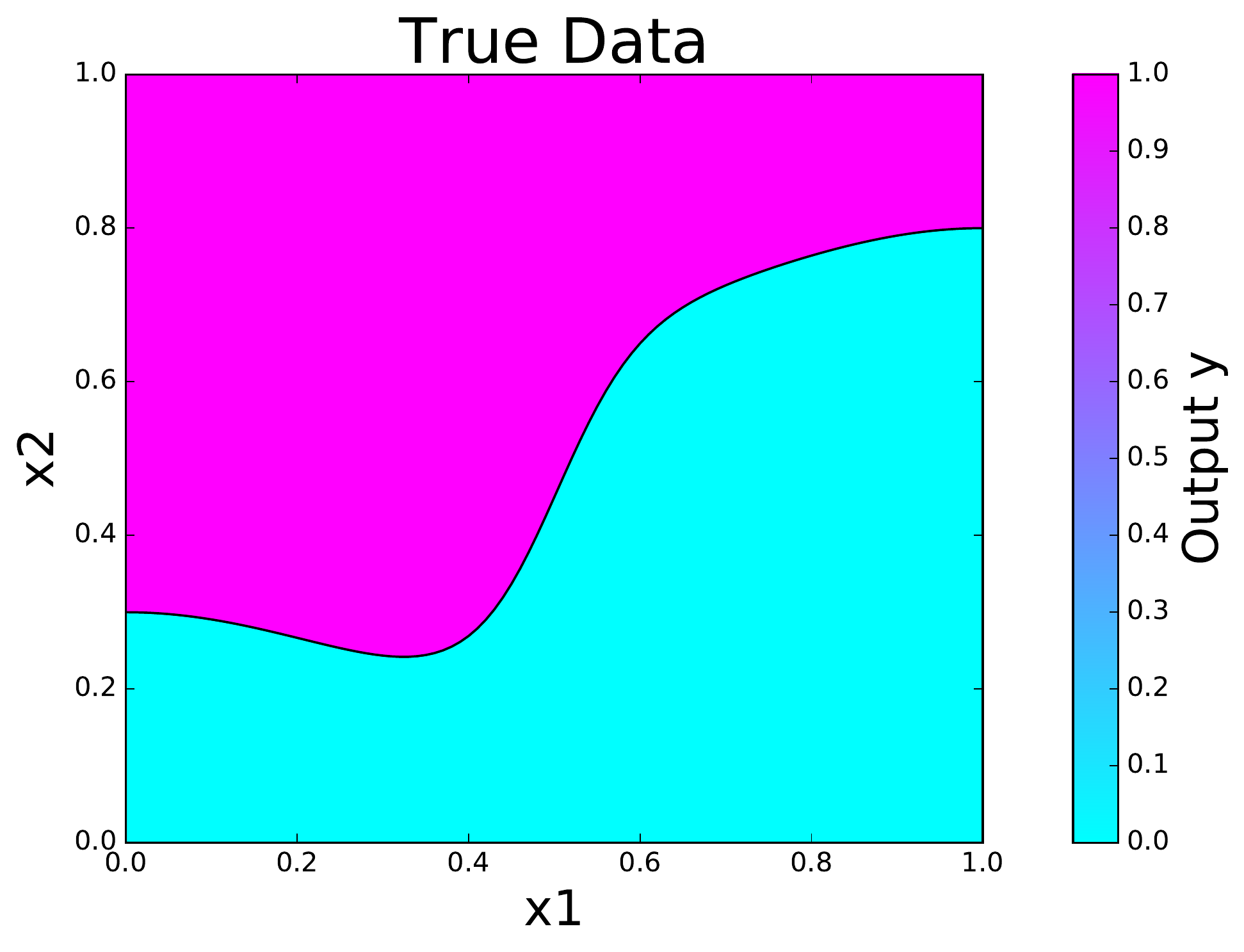}
		\end{minipage}
		\label{fig:true}}
	\subfigure[Learned Tree Ensembles]{
		\begin{minipage}[c]{0.31\textwidth}
		\includegraphics[width=0.95\textwidth]{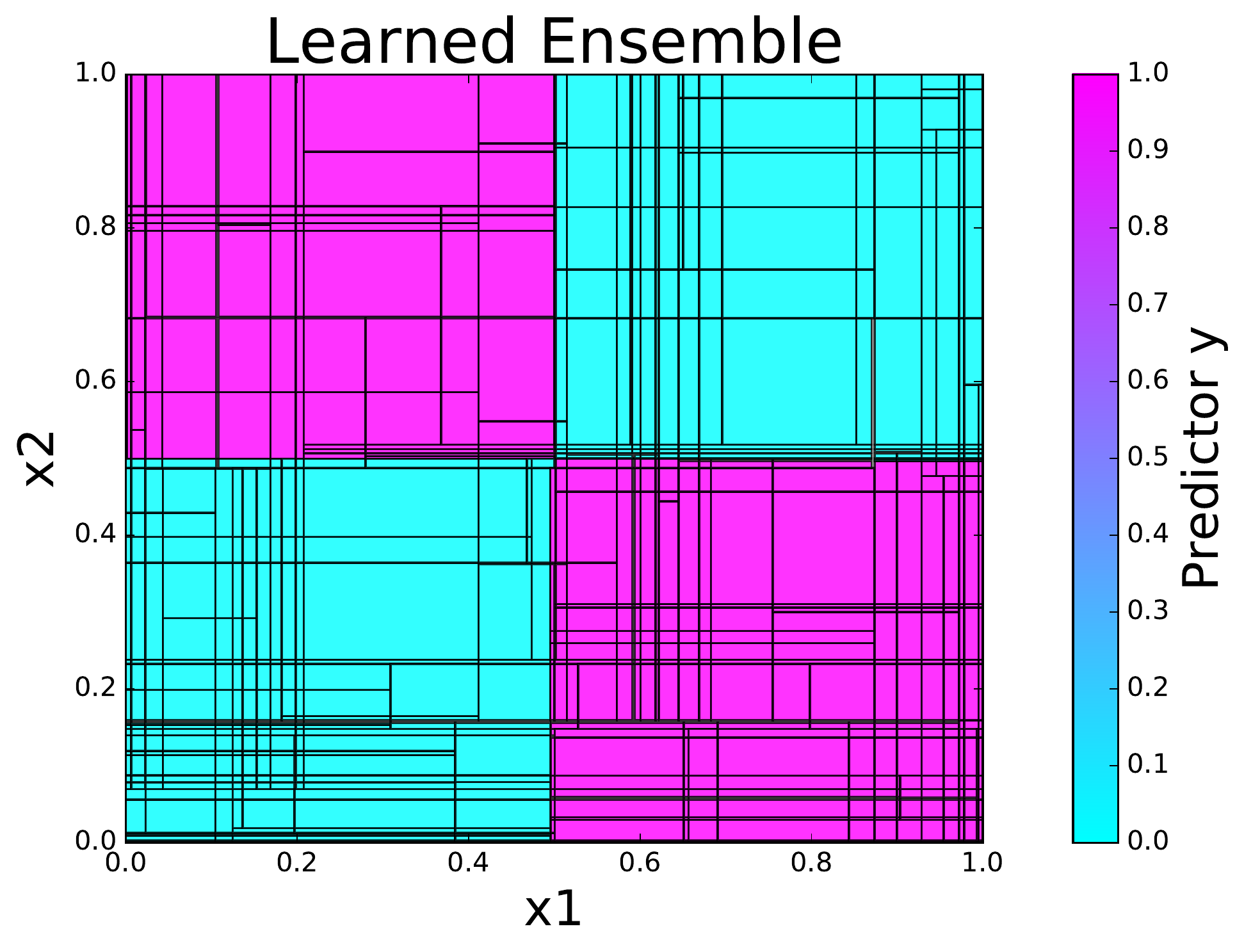}\\
		\includegraphics[width=0.95\textwidth]{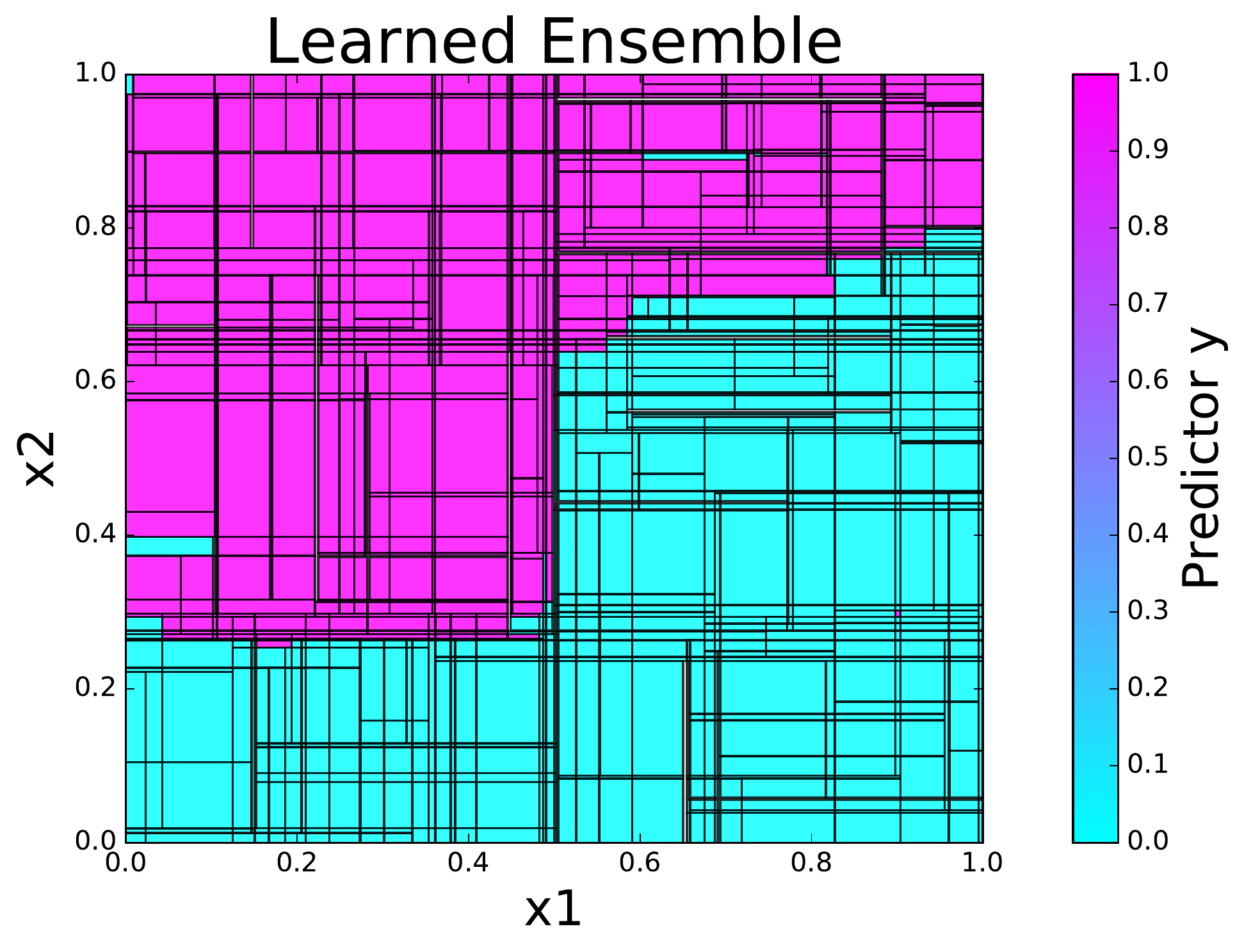}
		\end{minipage}
		\label{fig:before}}
	\subfigure[Simplified Models]{
		\begin{minipage}[c]{0.31\textwidth}
		\includegraphics[width=0.95\textwidth]{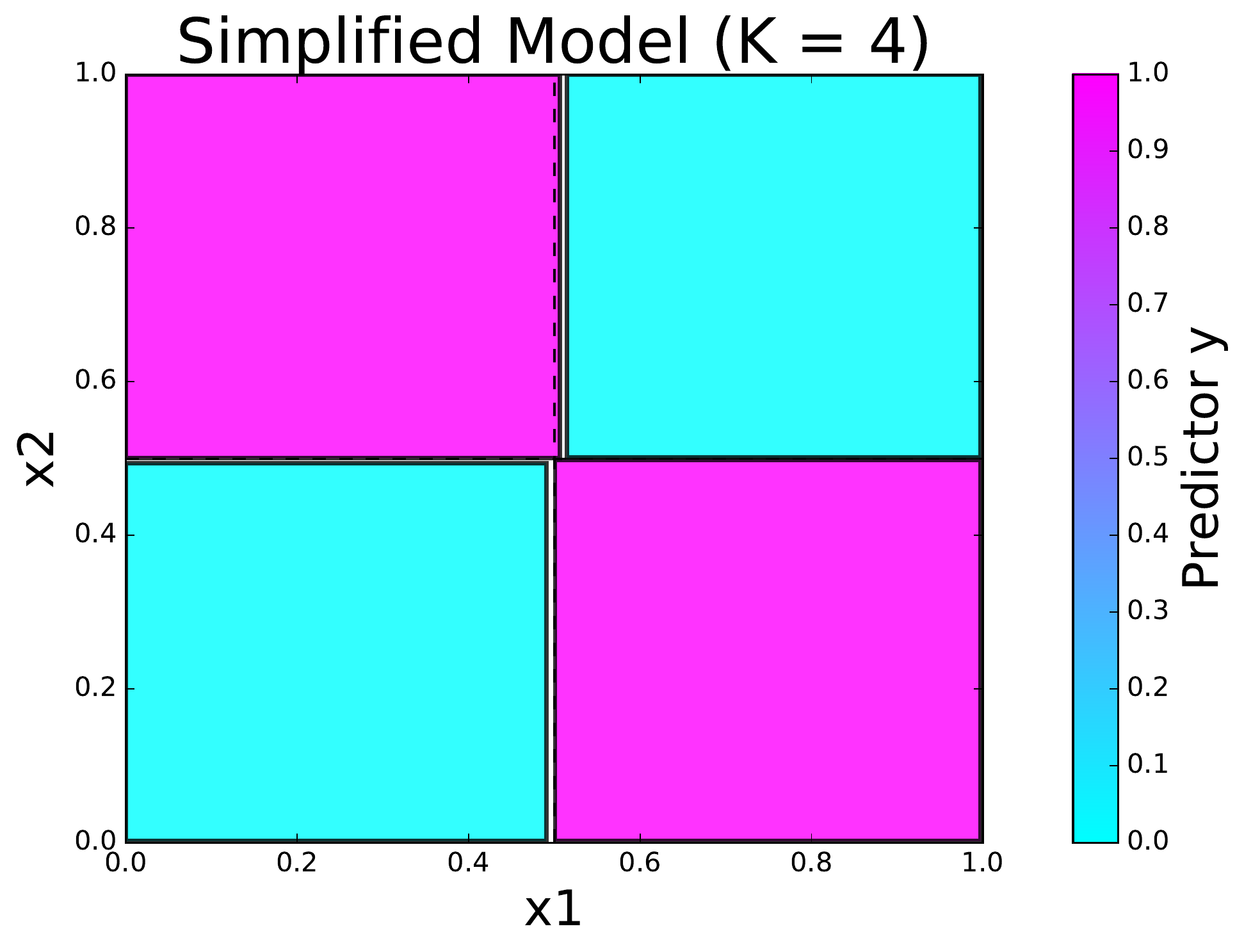}\\
		\includegraphics[width=0.95\textwidth]{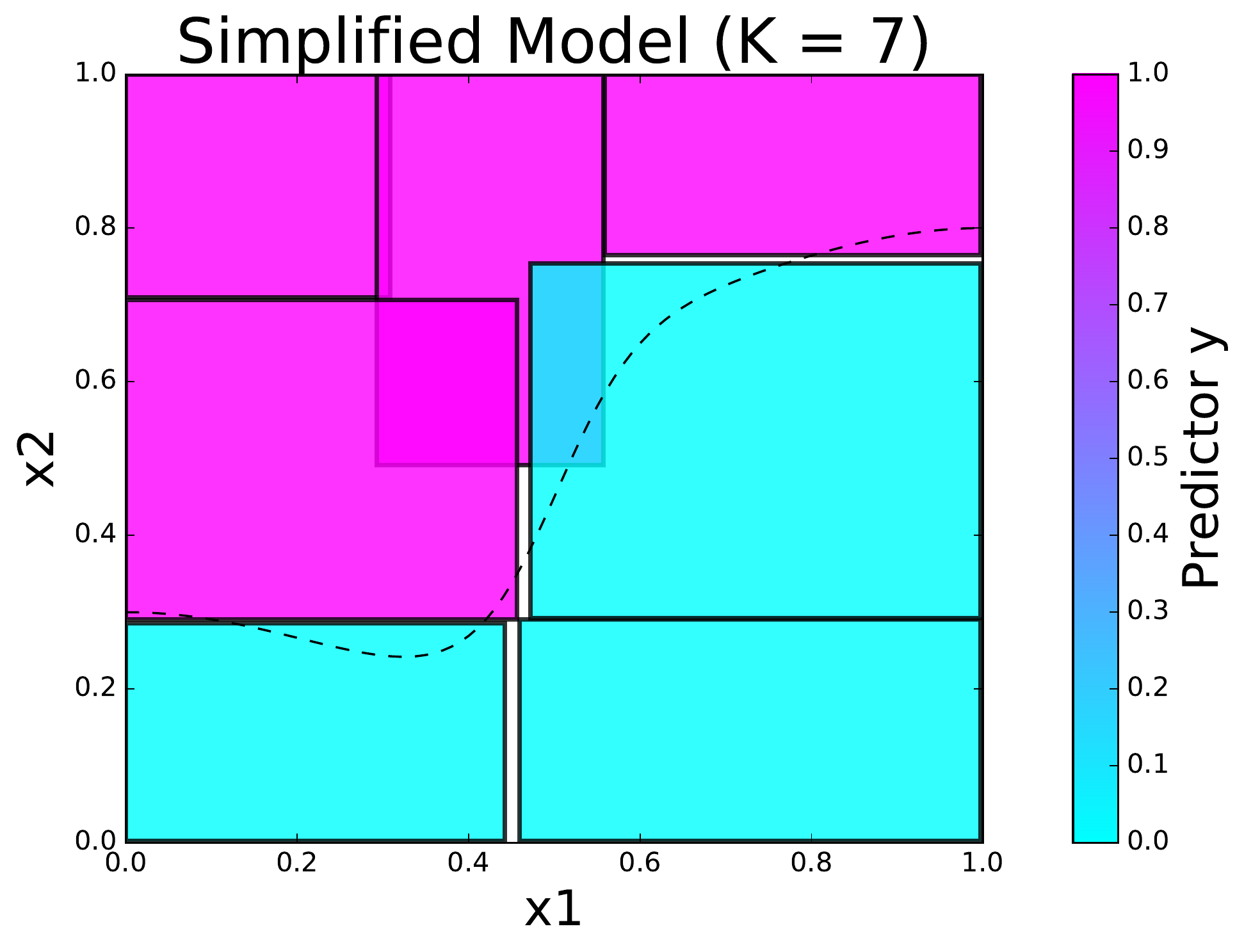} 
		\end{minipage}
		\label{fig:after}}
	\caption{The original data (a) are learned by tree ensembles with number of regions (b). In this example, the first five trees in the ensembles generated around 1,000 regions. The complicated ensembles (b) are \emph{defragged} into a few regions using the proposed method (c). Each rectangle shows each input region specified by the model.}
	\label{fig:example}
\end{figure*}

However, this high prediction performance of the tree ensemble makes large sacrifices of interpretability. 
Because every tree generates different regions, the resulting prediction model is inevitably \emph{fragmented}, i.e., it has a lot of redundancy and becomes considerably complicated (\figurename~\ref{fig:before}), even if the original data are simply structured (Figure~\ref{fig:true}).
The total number of regions is usually over a thousand, which roughly means that thousands of different rules are involved in the prediction.
Such a large number of rules are nearly impossible for humans to interpret.

How can we make a tree ensemble more interpretable?
Clearly, reducing the number of regions, or equivalently, reducing the number of rules, simplifies the model and improves its interpretability.
However, if the model is too simplified, we may overlook important rules behind the data.
Also, oversimplification of the model possibly degrades its prediction performance.
These observations imply that there is a trade-off between the number of regions and the prediction performance when simplifying the model.

In statistics, similar trade-offs have comprehensively been addressed as the model selection problem. Given multiple models, model selection methods typically aim to choose the model that achieves the best generalization performance, i.e., it can predict well for new data~\cite{akaike1974new,schwarz1978estimating}. Since too complex models cause over-fitting, simple models tend to be selected. 
One of the most popular model selection is Bayesian model selection~\cite{kassr95}. Bayesian model selection uses the marginal likelihood as a criterion, which eliminates redundant models as Occam's razor~\cite{schwarz1978estimating}. This is a desirable property for tree ensemble simplification---using Bayesian model selection, we can find a simplified expression of the tree ensemble with smaller number of regions that is essentially equivalent to the original one.

Though the Bayesian model selection is a promising approach, there are two difficulties. 
First, popular tree ensembles such as XGBoost are not probabilistic models, and their marginal likelihoods are not defined.
The Bayesian model selection is therefore not directly applicable to them. 
Second, the model simplification problem potentially incurs computational intractability.
For good model simplification, we have to make large regions that approximate the original tree ensemble as in \figurename~\ref{fig:after}.
However, because the possible configurations of input regions, e.g., the shape and the location of the regions, can be infinitely many, the full search of all possible candidates is infeasible.

In this study, we propose a simplification method for tree ensembles, both for classification and regression tasks, in a Bayesian manner.
Suppose that we are given a tree ensemble learned in a standard manner with a number of regions (Figure~\ref{fig:before}).
Our objective is to \emph{defrag} the tree ensemble into a simple model using a smaller number of regions, as shown in Figure~\ref{fig:after}.
Following the Bayesian principle, we tackle the first difficulty by adopting a probabilistic model representation of the tree ensemble.
For the second difficulty, we search for a good simple model by estimating a parameter of the probabilistic model.
With these modifications, the model simplification problem then reduces to the Bayesian model selection problem that optimizes input regions for simplification while maintaining the prediction performance as much as possible.
For efficient model selection, we adopt a Bayesian model selection algorithm called \emph{factorized asymptotic Bayesian (FAB) inference}~\cite{fujimaki2012factorized,hayashi2015rebuilding}.
FAB inference provides an asymptotic approximation of the marginal likelihood in a tractable form.
In addition, FAB inference has an automatic model pruning mechanism so that the costly outer loop for searching several possible model candidates is not necessary.
Our numerical experiments on several datasets showed that complicated tree ensembles were approximated adequately while maintaining prediction performance.

\paragraph{Notation:}
For $N \in \mathbb{N}$, $[N] = \{1, \ldots, N\}$ denotes the set of integers.
For a statement $a$, $\mathbb{I}(a)$ denotes the indicator of $a$, i.e., $\mathbb{I}(a) = 1$ if $a$ is true, and $\mathbb{I}(a) = 0$ if $a$ is false.
Let $\bm{x}= (x_1, x_2, ..., x_D)\in\mathbb{R}^D$ be a $D$-dimensional input and $y\in\mathcal{Y}$ be an output.
Here, for regression problems, the output $y$ is numeric, i.e., $\mathcal{Y}=\R$.
For classification problems with $C$ categories, the output $y$ is one-hot vector, i.e., for category $c \in [C]$, $y_c=1$ and $y_{c'} = 0$ for $c \neq c'$.

\section{Preliminaries}
\label{sec:tree}

\subsection{Decision Tree}
\label{sec:dtree}

The decision tree makes the predication depending on the leaf node to which the input $\bm{x}$ belongs.
The corresponding leaf node is determined by traversing the tree from the root.
In each internal node $j$ of the tree, the input $\bm{x}$ is directed to one of two child nodes depending on whether the statement $x_{d_j} > b_j$ is true or not, where $d_j\in[D]$ is a feature index checked at the node $j$ and $b_j\in\mathbb{R}$ is a threshold.
For example, suppose the case that $D=3$ and the leaf node $i$ is described by four internal nodes as $x_{1} > b_1$, $x_{2} \leq b_2$, $x_{3} > b_3$, $x_{3} \leq b'_3$, and let $\tilde{z}_i\in\mathcal{Y}$ be the predictive value of the leaf node $i$.
Then, if the input $\bm{x}$ arrives at the leaf node $i$ by traversing these internal nodes, the prediction mechanism is described as a \emph{rule}:
\begin{align*}
 	\underbrace{x_{1} > b_1}_{statement} \land \underbrace{x_{2} \leq b_2}_{statement} \land \underbrace{x_{3} > b_3}_{statement} \land \underbrace{x_{3} \leq b'_3}_{statement} \Longrightarrow y = \tilde{z}_i.
\end{align*}
We refer to each component as a \textit{statement} hereafter.

A list of statements can also be represented as a \textit{region}.
The above list of statements (the left hand side of the rule) can be written as $\bm{x} \in \tilde{R}_i := (b_1, \infty) \times (-\infty, b_2] \times (b_3, b'_3]$.
Note that the regions are mutually disjoint, i.e., $\tilde{R}_i \cap \tilde{R}_{i'} = \emptyset$ if $i \neq i'$.
Letting $\tilde{\mathcal{Z}}=\{\tilde{z}_i\}_{i=1}^I$ and $\tilde{\mathcal{R}}=\{\tilde{R}_i\}_{i=1}^I$, the decision tree  with $I$ leaf nodes can be expressed as follows:
\begin{align}
	f(\bm{x}; \tilde{\mathcal{Z}},\tilde{\mathcal{R}}, I) := 
	\sum_{i=1}^I \tilde{z}_i \mathbb{I}(\bm{x} \in \tilde{R}_i).
	\label{eq:dtree}
\end{align}

\subsection{Tree Ensemble}
\label{sec:ensemble}

The tree ensemble makes a prediction by combining the outputs from $T$ decision trees.
Suppose that the $t$-th decision tree has $I_t$ leaf nodes with predictive values $\tilde{\mathcal{Z}}_t$ and regions $\tilde{\mathcal{R}}_t$.
With weights $w_t \in \mathbb{R}$ on each decision tree $t \in [T]$, the output of the tree ensemble $y$ is determined by the weighted average $y = \sum_{t=1}^T w_t f(\bm{x};\tilde{\mathcal{Z}}_t,\tilde{\mathcal{R}}_t,I_t)$ for regression, or the weighted voting $y = \argmaxl_{c} \sum_{t=1}^T w_t \mathbb{I}(f(\bm{x};\tilde{\mathcal{Z}}_t,\tilde{\mathcal{R}}_t,I_t) = c)$ for classification.

\subsection{Extracting Rules from Tree Ensemble}
\label{sec:ensemble_rule}

To interpret the tree ensemble, we need to extract rules from it.
This corresponds to finding input regions and corresponding predictive values as in the single tree case~(\ref{eq:dtree}).
This can be achieved by considering multiple regions assigned by each tree to the input $\bm{x}$.
Suppose that the region $\tilde{R}_{i_t}^t$ is assigned to the input $\bm{x}$ in the $t$-th tree for each $t \in [T]$.
This means that the input $\bm{x}$ belongs to the intersection of those regions, namely $R_g = \cap_{t=1}^T \tilde{R}_{i_t}^t$.
The predictive value corresponding to $R_g$ can be expressed as $z_g = \sum_{t=1}^T w_t \tilde{z}_{i_t}^t$ for regression, and $z_g = \argmaxl_{c} \sum_{t=1}^T w_t \mathbb{I}(\tilde{z}_{i_t}^t = c)$ for classification.
As the result, the tree ensemble can be expressed as
\begin{align}
	f(\bm{x};\mathcal{Z},\mathcal{R},G) = \sum_{g=1}^G z_g \mathbb{I}(\bm{x} \in R_g),
	\label{eq:ensemble}
\end{align}
where $G$ is the number of total regions determined by the tree ensemble, $\mathcal{Z}=\{z_g\}_{g=1}^G$, and $\mathcal{R}=\{R_g\}_{g=1}^G$.
Because each region and predictive value corresponds to a rule, now we obtain the rules of the tree ensemble as $\mathcal{Z}$ and $\mathcal{R}$.

\section{Tree Ensemble Simplification Problem}
\label{sec:problem}

The expression (\ref{eq:ensemble}) indicates that the tree ensemble can be expressed using $G$ regions.
Because $\mathcal{R}$ is generated from all the possible combination of the regions of the individual trees $\tilde{\mathcal{R}}_1,\dots,\tilde{\mathcal{R}}_T$, the number of regions $G$ can grow exponentially in the number of trees $T$, which makes the interpretation of the tree ensemble almost impossible.
For example, \figurename~\ref{fig:before} shows that even five trees can generate more than a thousand regions.

To make the tree ensemble with large $G$ interpretable, we approximate it using a smaller number of regions.
Once the tree ensemble is approximated using a few regions as in \figurename~\ref{fig:after}, it is easy to interpret the underlying rules in the model.
This idea is formulated as the following problem.
\begin{problem}
Given $G\in\mathbb{N}$ predictive values $\mathcal{Z}=\{z_g\}_{g=1}^G$ and regions $\mathcal{R}=\{R_g\}_{g=1}^G$, find $K\ll G$ predictive values $\mathcal{Z}'=\{z'_k\}_{k=1}^K$ and regions $\mathcal{R}'=\{R'_k\}_{k=1}^K$ such that
\begin{align}
	f(\bm{x};\mathcal{Z},\mathcal{R},G) \approx f(\bm{x};\mathcal{Z}',\mathcal{R}',K)
	\label{eq:approx}
\end{align}
for any $\bm{x}\in\R^D$.
	\label{prob:approx}
\end{problem}

\section{Tree Ensemble as a Probabilistic Model}
\label{sec:prob}

To solve Problem~\ref{prob:approx}, we need to optimize the number of regions $K$, the predictors $\mathcal{Z}'$, and the regions $\mathcal{R}'$.
Here, we introduce a probabilistic model that expresses the predictive values and the regions.

\subsection{Binary Vector Expression of the Regions}
\label{sec:threshold}

\begin{figure}[tb]
	\centering
	\begin{tikzpicture}[scale=0.9]
		\draw[thick,->] (0.5,-0.5) -- (5,-0.5) node[right] {\large $x_1$};
		\draw[thick,->] (0.5,-0.5) -- (0.5,2.2) node[above] {\large $x_2$};
		\draw[dashed] (1, 2.2) -- (1, -0.5) node[below] {$b_1$};
		\draw[dashed] (3, 2.2) -- (3, -0.5) node[below] {$b_2$};
		\draw[dashed] (4, 2.2) -- (4, -0.5) node[below] {$b_3$};
		\draw[dashed] (4.8, 0) -- (0.5, 0) node[left] {$b_4$};
		\draw[dashed] (4.8, 0.8) -- (0.5, 0.8) node[left] {$b_5$};
		\draw[dashed] (4.8,1.4) -- (0.5, 1.4) node[left] {$b_6$};
		\draw[dashed] (4.8, 1.8) -- (0.5, 1.8) node[left] {$b_7$};
		\draw[pattern=north west lines, pattern color=gray] (1, 0) -- (3, 0) -- (3, 0.8) -- (1, 0.8) -- (1, 0);
		\node (R1) at (2.0, 0.4) {$R_g$};
		\draw[pattern=north west lines, pattern color=gray] (3, 0.8) -- (4, 0.8) -- (4, 1.4) -- (3, 1.4) -- (3, 0.8);
		\node (R2) at (3.5, 1.1) {$R_{g'}$};
      \draw[dotted] (R1.south) -- (2.0, -1) node[below]{$\tilde{\bm{\eta}}_g = (1, 0, 0, 1, 0, 0, 0)$};
		\draw[dotted] (R2.north) -- (3.5, 2.3) node[above]{$\tilde{\bm{\eta}}_{g'} = (1, 1, 0, 1, 1, 0, 0)$};
	\end{tikzpicture}
	\caption{Example of binary vector expression of regions. The region $R_g = \{(x_1, x_2) \mid b_1 < x_1 \leq b_2, b_4 < x_2 \leq b_5\}$ is expressed by the binary vector $\tilde{\bm{\eta}}_g = (1, 0, 0, 1, 0, 0, 0)$: the first element of $\tilde{\bm{\eta}}_g$ is 1 because $R_g$ satisfies $x_1 > b_1$ while the second element of $\tilde{\bm{\eta}}_g$ is 0 because $R_g$ does not satisfy $x_1 > b_2$. The region $R_{g'}$ is also expressed by $\tilde{\bm{\eta}}_{g'}$ in the similar manner.}
	\label{fig:binary}
\end{figure}
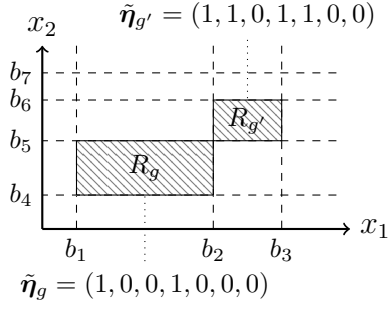

First, we modify the representation of the input $\bm{x}$ and the regions $\mathcal{R}$ for later convenience.
Suppose that the tree ensemble consists of $L$ statements in total, i.e.,  the decision trees in the ensemble have $L$ internal nodes in total.
By definition, each input region $R_g\in\mathcal{R}$ is uniquely characterized by the combination of $L$ statements, 
 meaning that $R_g$ is represented by the binary vector $\tilde{\bm{\eta}}_g \in \{0, 1\}^L$, where $\tilde{\eta}_{g \ell} = 1$ if $x_{d_{\ell}} > b_{\ell}$ for all $\bm{x}\in R_g$, and $\tilde{\eta}_{g \ell} = 0$ otherwise.
\figurename~\ref{fig:binary} illustrates an example.
By using this binary vector $\tilde{\bm{\eta}}_g$, the next equation holds:
\begin{align}
	\mathbb{I}(\bm{x} \in R_g) = \mathbb{I}(\bm{s}(\bm{x}) = \tilde{\bm{\eta}}_g) , 
	\label{eq:ensemble2}
\end{align}
where the $\ell$-th element of $\bm{s}(\bm{x})\in\{0,1\}^L$ is defined as $s_{\ell}(\bm{x}) = \mathbb{I}(x_{d_\ell} > b_{\ell})$.
To simplify the notation, we use $\bm{s}$ to denote $\bm{s}(\bm{x})$, and we refer to $\bm{s}$ as a \textit{binary feature} of the input $\bm{x}$.

\subsection{Probabilistic Model Expression of the Regions}
\label{sec:region}

As shown in \figurename~\ref{fig:before}, the tree ensemble splits the input region into small fragments. To derive a simplified model as in \figurename~\ref{fig:after}, we need to merge the small fragments to make a large region. We achieve this by interpreting the region $R_g$ as a generative model of the binary feature $\bm{s}$.

From the definition, the next equation holds:
\begin{align}
	\mathbb{I}(\bm{s} = \tilde{\bm{\eta}}_g) = \prod_{\ell=1}^L \tilde{\eta}_{g \ell}^{s_\ell} (1 - \tilde{\eta}_{g \ell})^{1 - s_{\ell}} .
	\label{eq:bernoulli}
\end{align}
The equation can be ``soften'' by extending $\tilde{\bm{\eta}}_g$ from a binary vector to a $[0,1]$-continuous vector $\bm{\eta}_g \in [0, 1]^L$.
Namely, now the right-hand-side of (\ref{eq:bernoulli}) is the Bernoulli distribution on $\bm{s}$ with a model parameter $\bm{\eta}_g$ where $\eta_{g \ell}$ indicates a probability $\eta_{g \ell} = p(s_{\ell} = 1) = p(x_{d_{\ell}} > b_{\ell})$.

\begin{figure}[tb]
	\centering
	\begin{tikzpicture}[scale=0.9]
		\draw[thick,->] (0.5,-0.5) -- (5,-0.5) node[right] {\large $x_1$};
		\draw[thick,->] (0.5,-0.5) -- (0.5,2.2) node[above] {\large $x_2$};
		\draw[dashed] (1, 2.2) -- (1, -0.5) node[below] {$b_1$};
		\draw[dashed] (3, 2.2) -- (3, -0.5) node[below] {$b_2$};
		\draw[dashed] (4, 2.2) -- (4, -0.5) node[below] {$b_3$};
		\draw[dashed] (4.8, 0) -- (0.5, 0) node[left] {$b_4$};
		\draw[dashed] (4.8, 0.8) -- (0.5, 0.8) node[left] {$b_5$};
		\draw[dashed] (4.8,1.4) -- (0.5, 1.4) node[left] {$b_6$};
		\draw[dashed] (4.8, 1.8) -- (0.5, 1.8) node[left] {$b_7$};
		\draw[pattern=north west lines, pattern color=gray] (1, 0) -- (4, 0) -- (4, 1.4) -- (1, 1.4) -- (1, 0);
		\node (Rg) at (2.5, 0.5) {$R_g$};
		\node[circle,draw=black, fill=white, inner sep=0pt,minimum size=5pt] (x1) at (1.5,0.4) {};
		\draw[dotted] (x1.south) -- (1.0, -1.0) node[below]{$\bm{s} = (1, 0, 0, 1, 0, 0, 0)$};
		\node[circle,draw=black, fill=white, inner sep=0pt,minimum size=5pt] (x2) at (3.5,0.4) {};
		\draw[dotted] (x2.south) -- (4.5, -1.0) node[below]{$\bm{s} = (1, 1, 0, 1, 0, 0, 0)$};
		\node[circle,draw=black, fill=white, inner sep=0pt,minimum size=5pt] (x3) at (1.5,1.1) {};
		\draw[dotted] (x3.north) -- (1.0, 2.5) node[above]{$\bm{s} = (1, 0, 0, 1, 1, 0, 0)$};
		\node[circle,draw=black, fill=white, inner sep=0pt,minimum size=5pt] (x4) at (3.5,1.1) {};
		\draw[dotted] (x4.north) -- (4.5, 2.5) node[above]{$\bm{s} = (1, 1, 0, 1, 1, 0, 0)$};
      \node (lr) at (2.5, -1.8) {$\Updownarrow$};
      \node (eg) at (2.5, -2.3) {$\bm{\eta}_g = (1, *, 0, 1, *, 0, 0)$};
	\end{tikzpicture}
	\caption{The concatenated region $R_g = \{(x_1, x_2) \mid b_1 < x_1 \leq b_3, b_4 < x_2 \leq b_6\}$ is indicated by the vector $\bm{\eta}_g = (1, *, 0, 1, *, 0, 0)$ where $*$ denotes the value between 0 and 1. This is because the region $R_g$ can be interpreted as a generative model of the binary feature $\bm{s}$. It can generate four different binary features each of which matches the pattern $\bm{\eta}_g = (1, *, 0, 1, *, 0, 0)$.}
	\label{fig:region}
\end{figure}
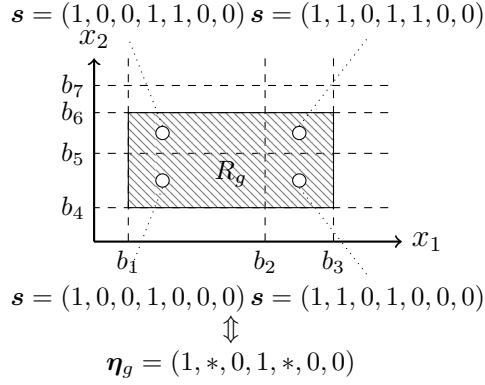

With the extended vector $\bm{\eta}_g$, we can now express the concatenated region using $\bm{\eta}_g$ as shown in \figurename~\ref{fig:region}.
Here, $\eta_{g \ell} = 1$ means that the region $R_g$ satisfies $x_{d_\ell} > b_{\ell}$ while $\eta_{g \ell} = 0$ means that $R_g$ satisfies $x_{d_\ell} \leq b_{\ell}$.
Moreover, with the extended vector $\bm{\eta}_g$, we have a third case when $\eta_{g \ell} \in (0, 1)$: this corresponds to  the case when some of $\bm{x} \in R_g$ satisfies $x_{d_{\ell}} > b_{\ell}$ while some other $\bm{x}' \in R_g$ satisfies $x'_{d_{\ell}} \leq b_{\ell}$, i.e., the boundary $x_{d_{\ell}} = b_{\ell}$ is inside $R_g$, and hence does not affect the definition of the region $R_g$.

The probabilistic version of \eqref{eq:bernoulli} is then given as the following generative model:
\begin{align}
	p(\bm{s} | g) := \prod_{\ell=1}^L \eta_{g \ell}^{s_\ell} (1 - \eta_{g \ell})^{1 - s_{\ell}} ,
	\label{eq:bernoulli2}
\end{align}
where $\eta_{g \ell} = p(s_{\ell} = 1) = p(x_{d_{\ell}} > b_{\ell})$.
Note that $\bm{\eta}_g$ is now a parameter of the model such that its zero--one pattern represents the shape of the concatenated region $R_g$.
Hence, by optimizing the parameter $\bm{\eta}_g$, we can optimize the shape of the region $R_g$.
The resulting parameter $\bm{\eta}_g$ is then translated to the corresponding statements describing the region $R_g$ from its zero--one pattern.

\subsection{Probabilistic Model Expression of the Tree Ensemble}
\label{sec:probabilistic}

We finally extend the tree ensemble into a probabilistic model.
For this purpose, we introduce an indicator $\bm{u} \in \{0, 1\}^G$ that describes which region the input $\bm{x}$ belongs to, i.e., if $\bm{x}$ belongs to the region $R_g$, $u_g = 1$ and $u_{g'} = 0$ for $g' \neq g$.
We then model the probability of the pair ($y$, $\bm{s}$) given the indicator $\bm{u}$ as
\begin{align}
	p(y, \bm{s} | \bm{u}, G) = \prod_{g=1}^G \left( p(y | g) p(\bm{s} | g) \right)^{u_g},
	\label{eq:pys}
\end{align}
where $p(y | g)$ is the probability that $y$ is output from the region $R_g$ and $p(\bm{s} | g)$ is defined in (\ref{eq:bernoulli2}).
Specifically, we adopt the next output model for $p(y | g)$:
\begin{align}
	p(y | g) := \begin{cases}
	\mathcal{N}(y | \mu_g, \lambda_g^{-1}) , & {\rm (regression)} , \\
	\prod_{c=1}^C \gamma_{gc}^{y_c} , &  {\rm (classification)} .
	\end{cases}
	\label{eq:py}
\end{align}
We denote the parameter of (\ref{eq:py}) by $\phi$ which is given by $\phi = \{\mu_g, \lambda_g\}_{g=1}^G$ for regression and $\phi = \{ \{\gamma_{gc}\}_{c=1}^C \}_{g=1}^G$ with $\gamma_{gc} \geq 0$ and $\sum_{c=1}^C \gamma_{gc}=1$ for classification.
We also model the probability of $\bm{u}$ by $p(\bm{u}) = \prod_{g=1}^G \alpha_g^{u_g}$ where $\alpha_g \geq 0$ and $\sum_{g=1}^G \alpha_g = 1$.
Here, $\alpha_g$ represents the probability of an event $u_g = 1$, i.e., the probability that input $\bm{x}$ belongs to the region $R_g$.
Therefore we write as $\alpha_g = p(g | \alpha)$.
Hence, the overall probabilistic expression of the tree ensemble can be expressed as follows:
\begin{align}
	p(y, \bm{s}, \bm{u} | \Pi, G) & = \prod_{g=1}^G \left( p(y | g, \phi) p(\bm{s} | g, \eta) \right)^{u_g} p(u_g | \alpha) \nonumber \\
	& = \prod_{g=1}^G \left( p(y | g, \phi) p(\bm{s} | g, \eta) p(g | \alpha) \right)^{u_g}
	\label{eq:pysu}
\end{align}
where we explicitly written down the model parameters $\phi$, $\eta$, and $\alpha$ for each component, and $\Pi$ is the set of all parameters $\Pi = \{\phi, \eta, \alpha\}$.

\subsection{Prediction}
\label{sec:pred}

From (\ref{eq:pysu}), we can naturally derive the posterior distribution of $y$ given the binary feature $\bm{s}$ using Bayes' rule.
In the prediction stage, we want to derive the output $y$ with the maximum posterior.
However, searching for the maximum posterior is computationally demanding, and we therefore propose using the next two-step MAP estimate:
\begin{align}
	\hat{g} &:= \argmaxl_{g} p(g | \bm{s}, \Pi, G) , \\
	\hat{y} &:= \argmaxl_{y} p(y | \hat{g}, \phi) . \label{eq:pred} 
\end{align}
We first find the region $\hat{g}$ with the maximum posterior $p(g | \bm{s}, \Pi, G)  \propto p(\bm{s} | g, \eta) p(g | \alpha)$.
Then, we output $\hat{y}$ with the maximum posterior given $\hat{g}$.

\section{Bayesian Model Selection Algorithm}
\label{sec:fab}

Using the probabilistic model~(\ref{eq:pysu}), Problem~\ref{prob:approx} is solved by estimating the model parameter $\Pi$ and the number of regions $K \ll G$ so that the model in (\ref{eq:pysu}) is adequately simplified.
If the number of regions $K$ is known and fixed, we can derive the optimal model parameter $\Pi$ of the simplified model using the maximum-likelihood estimation with the EM algorithm (see Appendix~\ref{sec:em}).
For an unknown $K$, we need to optimize it so that we can derive a simplified model with appropriate complexity.
From standard Bayesian theory~\cite{kassr95}, this model selection problem can be formulated as the maximization of marginal log-likelihood. 
To solve the problem, we employ factorized asymptotic Bayesian (FAB) inference~\cite{fujimaki2012factorized,hayashi2015rebuilding}, a Bayesian model selection algorithm that determines $\Pi$ and $K$ simultaneously.
Because the number of regions $K$ is automatically determined using FAB inference, we can avoid searching several possible values of $K$.
Hence, we can solve the model selection problem efficiently.

\subsection{FAB Inference}

With observations $\mathcal{D} = \{(y^{(n)}, \bm{s}^{(n)})\}_{n=1}^N$, we aim to determine the optimal number of regions $K$ by maximizing the marginal log-likelihood given by $\log p(\mathcal{D} | K) = \log \int p(\mathcal{D} | \Pi, K) p(\Pi) d\Pi$ .
Here, the likelihood is given as $p(\mathcal{D} | \Pi, K) = \prod_{n} p(y^{(n)}, \bm{s}^{(n)} | \Pi, K) = \prod_{n} \sum_{\bm{u}^{(n)}} p(y^{(n)}, \bm{s}^{(n)}, \bm{u}^{(n)} | \Pi, K)$.

Because the maximization of the marginal log-likelihood is intractable, we instead maximize the lower bound which is given by
\begin{align}
	& \sum_{n=1}^N \sum_{k=1}^K \mathbb{E}_{q(U)}[u_k^{(n)}] \log p(y^{(n)} | k, \phi) p(\bm{s}^{(n)} | k, \eta) p(k | \alpha) \nonumber \\
	& - \omega \sum_{k=1}^K \log \left( \sum_{n=1}^N \mathbb{E}_{q(U)} [u_k^{(n)}] + 1 \right) + H(q(U)) ,
   \label{eq:lb}
\end{align}
where $\omega = ({\rm dim} \phi / K + L + 1) / 2$, and $q(U)$ is the distribution of $U$.
The derivation of this lower bound can be observed in the supplementary material (Appendix \ref{sec:lb}).
The EM-like FAB inference algorithm is then formulated as an alternating maximization of the lower bound with respect to $q$ (E-step) and the parameter $\Pi$ (M-step).
See Algorithm~\ref{alg:fab} for the pseudo code.

\paragraph{[E-Step]}

In E-Step, we update the distribution $q(U)$ so that the lower bound in (\ref{eq:lb}) is maximized.
Let $\beta_k^{(n)} = \mathbb{E}_{q(U)}[u_k^{(n)}] = q(u_k^{(n)})$.
The optimal $\beta_k^{(n)}$ can be derived by iterating the next update until convergence:
\begin{align}
	\beta_k^{(n)} \propto f_k^{(n)} \exp\left( - \frac{\omega}{\sum_{n=1}^N \beta_k^{(n)} + 1} \right) ,
	\label{eq:estep}
\end{align}
where $f_k^{(n)} = p(y^{(n)} | k, \phi) p(\bm{s}^{(n)} | k, \eta) p(k | \alpha)$.
See Appendix~\ref{sec:derivation} for the derivation of this update.

\paragraph{[M-Step]}

In M-Step, we update the parameter $\Pi$ so that the lower bound in (\ref{eq:lb}) is maximized.
Let $\beta_k^{(n)} = q(u_k^{(n)})$.
The parameter $\Pi = \{\phi, \eta, \alpha\}$ is then updated as
\begin{align}
	\begin{split}
	& \text{(regression):}
	\begin{cases}
    	& \mu_k = \frac{\sum_{n=1}^N \beta_k^{(n)} y^{(n)}}{\sum_{n=1}^N \beta_k^{(n)}}, \\
		& \lambda_k = \frac{\sum_{n=1}^N \beta_k^{(n)}}{\sum_{n=1}^N \beta_k^{(n)} (y^{(n)} - \mu_k)^2} ,
   \end{cases} \\
   & \text{(classification):} \;\;\; \gamma_{kc} = \frac{\sum_{n=1}^N \beta_k^{(n)} y_c^{(n)}}{\sum_{n=1}^N \beta_k^{(n)}}, \\
	& \eta_{k \ell} = \frac{\sum_{n=1}^N \beta_k^{(n)} s_{\ell}^{(n)}}{\sum_{n=1}^N \beta_k^{(n)}} , \qquad \alpha_k = \frac{1}{N} \sum_{n=1}^N \beta_k^{(n)} .
    \end{split}
    \label{eq:mstep}
\end{align}

\begin{figure}[t]
\centering
\begin{algorithm}[H]
	\caption{FAB Inference}
	\label{alg:fab}
	\begin{algorithmic}
		\REQUIRE Training data $\mathcal{D} = \{(y^{(n)}, \bm{s}^{(n)})\}_{n=1}^N$, maximum number of regions $K_{\max}$, tolerance $\delta$
		\ENSURE \# of regions $K$, Parameter $\Pi = \{\phi, \eta, \alpha\}$
		\STATE Initialize parameter $\Pi$ and $\{ \{q(u_k^{(n)})\}_{k=1}^{K_{\max}} \}_{n=1}^N$ randomly
		\STATE $K \leftarrow K_{\max}$
		\WHILE{lower bound not converged}
			\WHILE{not converged}
				\STATE Update $\{ \{q(u_k^{(n)})\}_{k=1}^K \}_{n=1}^N$ by (\ref{eq:estep})
			\ENDWHILE
			\STATE Remove $k$-th region when $\frac{1}{N}\sum_{n=1}^N q(u_k^{(n)}) < \delta$
			\STATE $K \leftarrow$ \# of active regions
			\STATE Update $\Pi$ by (\ref{eq:mstep})
		\ENDWHILE
	\end{algorithmic}
\end{algorithm}
\end{figure}

\paragraph{[Region Truncation]}
The iterative update of E-Step in (\ref{eq:estep}) induces truncation of the region~\cite{fujimaki2012factorized,hayashi2015rebuilding}.
For instance, when $\sum_{n=1}^N \beta_k^{(n)} = \epsilon \ll N$, in (\ref{eq:estep}),  the updated value $\beta_k^{(n)}$ is multiplied by $\exp ( - \omega / (\epsilon + 1)) \ll 1$ for all $n \in [N]$.
The iterative multiplication of this small value results in $q(u_k^{(n)}) \approx 0$ for all $n \in [N]$, which means that the $k$-th region can be removed without affecting the marginal log-likelihood.
With this region truncation, FAB inference automatically decides the number of regions $K$ within the iterative optimization.
Hence, we only need to specify a sufficiently large $K_{\rm max}$ as the initial value of $K$.
We note that we can leave $K_{\rm max}$ as a constant (say, $K_{\rm max} = 10$) rather than the tuning parameter.

\subsection{Solution Selection}

Because the objective function (\ref{eq:lb}) is non-concave, we may solve the maximization problem multiple times for several different initial parameters to obtain better solutions.
We then select the model that best represents the data.
Suppose that we have $M$ candidates $\{\Pi_m\}_{m=1}^M$.
We propose to select the parameter with the smallest training error
$\tilde{\Pi} = \argminl_{\Pi_m} {\rm Error}(\mathcal{D}, \Pi_m)$,
where 
\begin{align*}
	{\rm Error}(\mathcal{D}, \Pi) := 
	\begin{cases}
	\sum_{n=1}^N \left( y^{(n)} - \hat{y}^{(n)} \right)^2 , & \text{(regression)}, \\
	\sum_{n=1}^N \mathbb{I}(y^{(n)} \neq \hat{y}^{(n)}) ,  & \text{(classification)} ,
	\end{cases}
\end{align*}
and $\hat{y}^{(n)}$ is determined from (\ref{eq:pred}).

\subsection{Computational Complexity of FAB Inference}
\label{sec:complexity}

The time complexity of the proposed FAB inference is dominated by E-Step which is $O(K_{\max} L N + \zeta K_{\max} N)$, where $\zeta$ is the number of iterations in E-Step.
In E-Step, we first need to compute $f_k^{(n)}$ for all $k \in [K_{\max}]$ and $n \in [N]$ which requires $O(K_{\max} L N)$ time complexity.
We then iteratively update the value of $\beta_k^{(n)}$ based on (\ref{eq:estep}).
The one update step (\ref{eq:estep}) for all $k \in [K_{\max}]$ and $n \in [N]$ requires $O(K_{\max} N)$ time complexity.
The overall time complexity of E-Step is therefore $O(K_{\max} L N + \zeta K_{\max} N)$.
In M-Step, the update of $\phi$, $\eta$, and $\alpha$ require $O(K_{\max} N)$, $O(K_{\max} L N)$, and $O(K_{\max} N)$ time complexities, respectively.
These complexities are dominated by the complexity of E-Step, and thus can be ignored.

\section{Related Work}
\label{sec:rel}

Interpretability of complex machine learning models are now in high demand~\cite{kim2016proceedings,wilson2016proceedings}.
Interpreting learned models allows us to understand the data and predictions more deeply~\cite{ribeiro2016should,kim2016examples}, which may lead to effective usage of data and models. For instance, we may be able to design a better prediction model by fixing the \textit{bug}~\cite{kaufman2012leakage} in the model~\cite{lloyd2015statistical}, or we can make a better decision based on the insights on the model~\cite{kim2015mind,kim2016examples}.

There are a few seminal studies on interpreting tree ensembles, including \textit{Born Again Trees} (BATrees)~\cite{breiman1996born}, \textit{interpretable Trees} (inTrees)~\cite{deng2014interpreting}, and \textit{Node Harvest}~\cite{meinshausen2010node}.

\citet{breiman1996born} proposed building a single decision tree that mimics the tree ensemble.
In BATrees, the tree ensemble is used to generate additional samples that are used to find the best split in the tree node.
The single decision tree is then built to perform in a manner similar to that of the original tree ensemble.
An important note regarding BATree is that it tends to generate a deep tree, i.e., a tree with several complicated prediction rules which may be difficult to interpret.

The inTrees framework extracts rules from tree ensembles by treating tradeoffs among the frequency of the rules appearing in the trees, the errors made by the predictions, and the length of the rules.
The fundamental difficulty with inTrees is that its target is limited to the classification tree ensembles.
Regression tree ensembles are first transformed into the classification ones by discretizing the output, and then inTrees is applied to extract the rules.
The number of discretization levels remains as a tuning parameter, which severely affects the resulting rules.

Node Harvest simplifies tree ensembles by using the shallow parts of the trees.
In the first step, the shallow part of the trees (e.g., depth two trees) are extracted and the remaining parts are discarded.
Node Harvest then combines the shallow trees so that they fit the training data well.
The advantage of Node Harvest is that the combination stage can be formulated as a convex quadratic programming, and thus, the global optimal solution can be derived efficiently.
However, the shortcoming of Node Harvest is that the derived simplified model is still an ensemble of the shallow trees.
It is therefore still challenging to interpret the resulting simplified ensemble.
It is also important to note that Node Harvest is designed for regression.
Although it can handle binary classification as an extension of regression, it cannot handle classification with more than two categories.

Our proposed method overcomes the limitations of these existing methods; the resulting model tends to have only a few rules that are easy to interpret; it can handle both classification and regression tree ensembles; and there are no tuning parameters.

\section{Experiments}
\label{sec:exp}

\begin{table}[t]
	\caption{[Datasets] Four real world datasets are obtained from the UCI Machine Learning Repository~\cite{Lichman:2013}. The task of the first five data are binary classification, while the task of the last Energy data is regression. $D$ is the data dimensionality, $N_{\rm all}$ is the number of data points in the original dataset, and $N_{\rm train}$ and $N_{\rm test}$ denote the number of data points randomly sampled for training and testing the models in the experiment.}
	\label{tab:data}
	\centering
	\begin{tabular}{cccc}
		& $D$ & $N_{\rm all}$ & $N_{\rm train}, N_{\rm test}$ \\
		\hline
		Synthetic1 & 2 & - & 1,000 \\
		Synthetic2 & 2 & - & 1,000 \\
		Spambase & 57 & 4,601 & 1,000 \\
		MiniBooNE & 50 & 130,065 & 5,000 \\
		Higgs & 28 & 11,000,000 & 5,000 \\
		Energy & 8 & 768 & 384
	\end{tabular}
\end{table}

We demonstrate the efficacy of the proposed method through synthetic and real-world data applications~\footnote{The experiment codes are available at \url{https://github.com/sato9hara/defragTrees}.}.
The used data in this experiment are summarized in \tablename~\ref{tab:data}.
The first synthetic data (Synthetic1) is generated from the following procedure:
\begin{align*}
	\bm{x} &= (x_1, x_2) \sim {\rm Uniform}[0, 1],  \\
	y^* &= {\rm XOR}(x_1 > 0.5, x_2 > 0.5), \\
	y &= {\rm XOR}(y^*, \theta), 
\end{align*}
where $\theta \in \{0, 1\}$ with $p(\theta = 1) = 0.1$, which corresponds to the 10\% label noise.
Similarly, the second synthetic data (Synthetic2) is generated by replacing the second step with
\begin{align*}
	y^* &= \mathbb{I}(x_2 > r(x_1)), \\
	r(x_1) &= 0.25 + \frac{0.5}{1 + \exp(- 20(x_1 - 0.5))} + 0.05\cos(2 \pi x_1) .
\end{align*}
Synthetic1 has a box-shaped class boundary (upper figure of \figurename~\ref{fig:true}), and can be expressed by the region-based model using four regions.
On the other hand, Synthetic2 has a more complicated class boundary (bottom figure of \figurename~\ref{fig:true}).
Hence, it is more difficult to simplify the tree ensemble and derive a good approximate model.

\paragraph{Baseline Methods: }
We compared the proposed method to four baseline methods. 
The first three are the tree ensemble simplification methods: BATrees~\cite{breiman1996born}, inTrees~\cite{deng2014interpreting}, and Node Harvest (NH)~\cite{meinshausen2010node}.
The last baseline is the depth-2 decision tree (DTree2).
While the above three methods tend to generate tens or hundreds of rules, DTree2 generates only four rules.
Hence, it is a good baseline method to compare with the proposed method that tends to generate only a few rules.

\paragraph{Implementations: }
In all experiments, we used \texttt{randomForest} package in R to train tree ensembles with 100 trees.
The tree ensemble simplification methods are then applied to extract rules from the learned tree ensembles.
The proposed method is implemented in Python.
In the experiments, we set $K_{\max}=10$ and ran FAB inference for 20 different random initial parameters, and adopted the result with the smallest training error.
The BATrees is implemented also in Python.
The depth of BATrees is chosen from $\{2, 3, 4, 6, 8, 10\}$ using 5-fold cross validation.
For inTrees and Node Harvest, we used their R implementations with their default settings.
For DTree2, we used \texttt{DecisionTreeRegressor} and \texttt{DecisionTreeClassifier} of scikit-learn in Python while fixing their depth to two.
All experiments were conducted on 64-bit CentOS 6.7 with an Intel Xeon E5-2670 2.6GHz CPU and 512GB RAM.

\subsection{FAB Inference vs. EM Algorithm}

We compared the runtimes of FAB inference and the EM algorithm.
For the EM algorithm, we ran the algorithm by varying the value of $K$ from 1 to 10, and reported the total runtime.

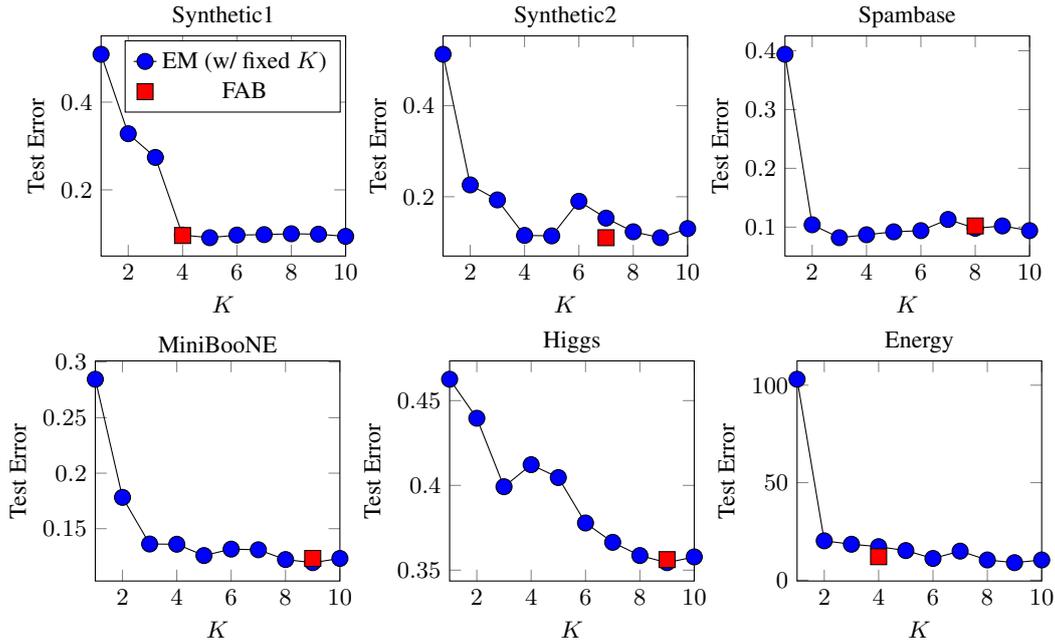
\begin{figure*}[t]
\small
\centering
\tikzset{mark options={mark size=3}}
\pgfplotsset{
compat=1.11,
legend image code/.code={
\draw[mark repeat=2,mark phase=2]
plot coordinates {
(0cm,0cm)
(0.15cm,0cm)        
(0.3cm,0cm)         
};%
}
}
\begin{tikzpicture} 
\begin{axis}[
	scale=1.0,
	xlabel={$K$},
	xmin=1,
	xmax=10,
	ylabel style={align=center},
	ylabel={Test Error},
	title style={yshift=-1.5ex,},
	title={Synthetic1},
	width=0.32\textwidth,
	height=4.5cm,
]
\addplot[
	scatter,scatter src=explicit symbolic,
	scatter/classes={
		EM={mark=*,draw=black,fill=blue}
		}
	]
	table [x=K, y=error, meta=label, col sep=comma,skip coords between index={10}{11}]{compare_synthetic4.csv};
\addplot[
	scatter,only marks,scatter src=explicit symbolic,
	scatter/classes={
		FAB={mark=square*,draw=black,fill=red}
		}
	]
	table [x=K, y=error, meta=label, col sep=comma,skip coords between index={0}{10}]{compare_synthetic4.csv};
	\legend{EM (w/ fixed $K$), FAB}
\end{axis}
\end{tikzpicture}
\hspace{-6pt}
\begin{tikzpicture} 
\begin{axis}[
	scale=1.0,
	xlabel={$K$},
	xmin=1,
	xmax=10,
	ylabel style={align=center},
	ylabel={Test Error},
	title style={yshift=-1.5ex,},
	title={Synthetic2},
	width=0.32\textwidth,
	height=4.5cm,
]
\addplot[
	scatter,scatter src=explicit symbolic,
	scatter/classes={
		EM={mark=*,draw=black,fill=blue}
		}
	]
	table [x=K, y=error, meta=label, col sep=comma,skip coords between index={10}{11}]{compare_synthetic3.csv};
\addplot[
	scatter,only marks,scatter src=explicit symbolic,
	scatter/classes={
		FAB={mark=square*,draw=black,fill=red}
		}
	]
	table [x=K, y=error, meta=label, col sep=comma,skip coords between index={0}{10}]{compare_synthetic3.csv};
\end{axis}
\end{tikzpicture}
\hspace{-6pt}
\begin{tikzpicture} 
\begin{axis}[
	scale=1.0,
	xlabel={$K$},
	xmin=1,
	xmax=10,
	ylabel style={align=center},
	ylabel={Test Error},
	title style={yshift=-1.5ex,},
	title={Spambase},
	width=0.32\textwidth,
	height=4.5cm,
]
\addplot[
	scatter,scatter src=explicit symbolic,
	scatter/classes={
		EM={mark=*,draw=black,fill=blue}
		}
	]
	table [x=K, y=error, meta=label, col sep=comma,skip coords between index={10}{11}]{compare_spambase.csv};
\addplot[
	scatter,only marks,scatter src=explicit symbolic,
	scatter/classes={
		FAB={mark=square*,draw=black,fill=red}
		}
	]
	table [x=K, y=error, meta=label, col sep=comma,skip coords between index={0}{10}]{compare_spambase.csv};
\end{axis}
\end{tikzpicture}

\begin{tikzpicture} 
\begin{axis}[
	scale=1.0,
	xlabel={$K$},
	xmin=1,
	xmax=10,
	ylabel style={align=center},
	ylabel={Test Error},
	title style={yshift=-1.5ex,},
	title={MiniBooNE},
	width=0.32\textwidth,
	height=4.5cm,
]
\addplot[
	scatter,scatter src=explicit symbolic,
	scatter/classes={
		EM={mark=*,draw=black,fill=blue}
		}
	]
	table [x=K, y=error, meta=label, col sep=comma,skip coords between index={10}{11}]{compare_miniboone.csv};
\addplot[
	scatter,only marks,scatter src=explicit symbolic,
	scatter/classes={
		FAB={mark=square*,draw=black,fill=red}
		}
	]
	table [x=K, y=error, meta=label, col sep=comma,skip coords between index={0}{10}]{compare_miniboone.csv};
\end{axis}
\end{tikzpicture}
\hspace{-6pt}
\begin{tikzpicture} 
\begin{axis}[
	scale=1.0,
	xlabel={$K$},
	xmin=1,
	xmax=10,
	ylabel style={align=center},
	ylabel={Test Error},
	title style={yshift=-1.5ex,},
	title={Higgs},
	width=0.32\textwidth,
	height=4.5cm,
]
\addplot[
	scatter,scatter src=explicit symbolic,
	scatter/classes={
		EM={mark=*,draw=black,fill=blue}
		}
	]
	table [x=K, y=error, meta=label, col sep=comma,skip coords between index={10}{11}]{compare_higgs.csv};
\addplot[
	scatter,only marks,scatter src=explicit symbolic,
	scatter/classes={
		FAB={mark=square*,draw=black,fill=red}
		}
	]
	table [x=K, y=error, meta=label, col sep=comma,skip coords between index={0}{10}]{compare_higgs.csv};
\end{axis}
\end{tikzpicture}
\hspace{-6pt}
\begin{tikzpicture} 
\begin{axis}[
	scale=1.0,
	xlabel={$K$},
	xmin=1,
	xmax=10,
	ylabel style={align=center},
	ylabel={Test Error},
	title style={yshift=-1.5ex,},
	title={Energy},
	width=0.32\textwidth,
	height=4.5cm,
]
\addplot[
	scatter,scatter src=explicit symbolic,
	scatter/classes={
		EM={mark=*,draw=black,fill=blue}
		}
	]
	table [x=K, y=error, meta=label, col sep=comma,skip coords between index={10}{11}]{compare_energy.csv};
\addplot[
	scatter,only marks,scatter src=explicit symbolic,
	scatter/classes={
		FAB={mark=square*,draw=black,fill=red}
		}
	]
	table [x=K, y=error, meta=label, col sep=comma,skip coords between index={0}{10}]{compare_energy.csv};
\end{axis}
\end{tikzpicture}
\caption{Test Errors of FAB inference and the EM algorithm.}
\label{fig:compare}
\end{figure*}

\tablename~\ref{tab:time} summarizes that FAB inference was from 5 to 20 times faster than the EM algorithm.
FAB inference attained smaller runtimes by avoiding searching over several possible number of rules $K$ and deciding the number automatically.
\figurename~\ref{fig:compare} shows the comparison of the test errors of the found rules: they show that FAB inference could find an appropriate number of rules $K$ with small prediction errors.
These results suggest the superiority of FAB inference over the EM algorithm as it could find an appropriate number of rules by avoiding redundant computations for searching the number of rules $K$.

\begin{table*}[t]
	\centering
   \caption{Average runtimes in seconds for one restart: the EM algorithm ran over $K=1, 2, \ldots, 10$, and its total time is reported.}
   \label{tab:time}
	\begin{tabular}{ccccccc}
    & Synthetic1 & Synthetic2 & Spambase & MiniBooNE & Higgs & Energy \\
    \hline
   	FAB & $2.59 \pm 2.28$& $4.03 \pm 4.76$ & $3.07 \pm 1.35$ & $280 \pm 174$ & $149 \pm 58.3$ &$0.03 \pm 0.02$  \\ 
    EM & $21.4 \pm 7.33$ & $30.1 \pm 11.4$ & $81.2 \pm 12.4$ & $1459 \pm 263$ & $603 \pm 53.6$ & $0.49 \pm 0.18$ \\
	\end{tabular}
\end{table*}

\subsection{Comparison with the Baseline Methods}

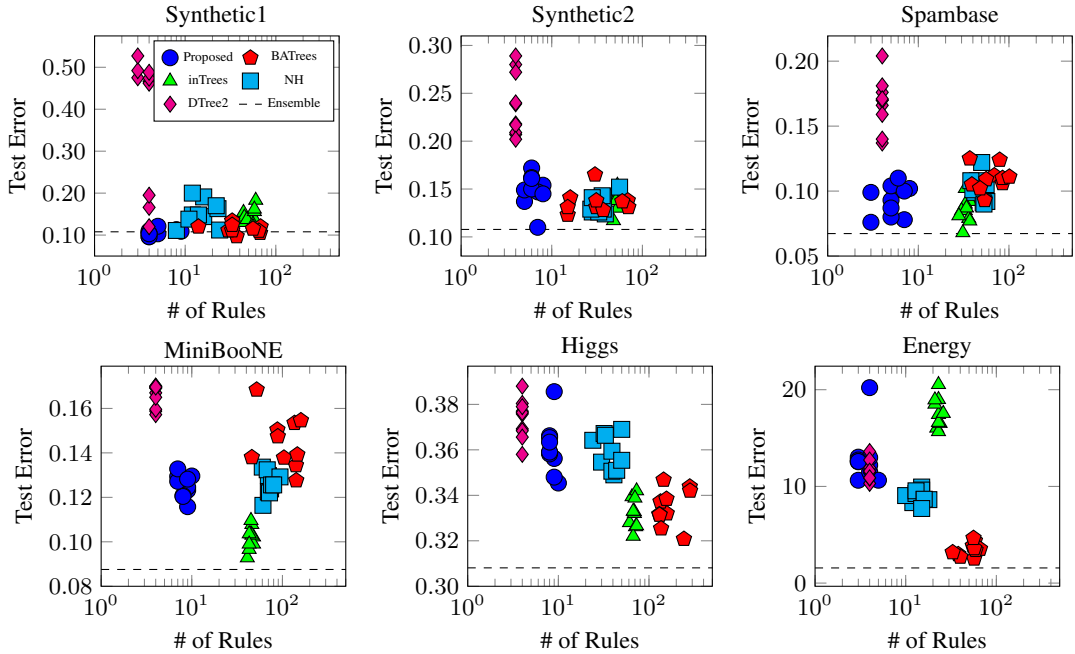
\begin{figure*}[t]
\small
\centering
\tikzset{mark options={mark size=3}}
\pgfplotsset{
compat=1.11,
legend image code/.code={
\draw[mark repeat=2,mark phase=2]
plot coordinates {
(0cm,0cm)
(0.15cm,0cm)        
(0.3cm,0cm)         
};%
}
}
\begin{tikzpicture} 
\begin{semilogxaxis}[
	scale=1.0,
	xlabel={\# of Rules},
	xmin=1,
	xmax=500,
	ylabel style={align=center},
	ylabel={Test Error},
	ymin=0.05,
	ytick={0.1, 0.2, 0.3, 0.4, 0.5},
	yticklabel style={
		/pgf/number format/fixed,
		/pgf/number format/precision=2,
		/pgf/number format/fixed zerofill
	},
	title style={yshift=-1.5ex,},
	title={Synthetic1},
	width=0.32\textwidth,
	height=4.5cm,
	legend style={font=\fontsize{5}{5}\selectfont},
	legend columns=2,
]
\addplot[
	scatter,only marks,scatter src=explicit symbolic,
	scatter/classes={
		defrag={mark=*,draw=black,fill=blue},
		BATree={mark=pentagon*,draw=black,fill=red},
		inTreesFull={mark=triangle*,draw=black,fill=green},
		NH={mark=square*,draw=black,fill=cyan},
		DTree2={mark=diamond*,draw=black,fill=magenta}
		}
	]
	table [x=rule, y=error, meta=label, col sep=comma,skip coords between index={50}{52}]{csv_synthetic4.csv};
	\addplot[mark=none, color=black, dashed] table [y=error, meta=label, col sep=comma,skip coords between index={0}{50}]{csv_synthetic4.csv};
	\legend{Proposed,BATrees,inTrees,NH,DTree2, Ensemble}
\end{semilogxaxis}
\end{tikzpicture}
\hspace{6pt}
\begin{tikzpicture} 
\begin{semilogxaxis}[
	scale=1.0,
	xlabel={\# of Rules},
	xmin=1,
	xmax=500,
	ylabel style={align=center},
	ylabel={Test Error},
	ymin=0.08,
	ytick={0.1, 0.15, 0.2, 0.25, 0.3},
	yticklabel style={
		/pgf/number format/fixed,
		/pgf/number format/precision=2,
		/pgf/number format/fixed zerofill
	},
	title style={yshift=-1.5ex,},
	title={Synthetic2},
	width=0.32\textwidth,
	height=4.5cm,
    legend style={font=\fontsize{6}{5}\selectfont},
	legend columns=2,
]
\addplot[
	scatter,only marks,scatter src=explicit symbolic,
	scatter/classes={
		defrag={mark=*,draw=black,fill=blue},
		BATree={mark=pentagon*,draw=black,fill=red},
		inTreesFull={mark=triangle*,draw=black,fill=green},
		NH={mark=square*,draw=black,fill=cyan},
		DTree2={mark=diamond*,draw=black,fill=magenta}
		}
	]
	table [x=rule, y=error, meta=label, col sep=comma,skip coords between index={50}{52}]{csv_synthetic3.csv};
	\addplot[mark=none, color=black, dashed] table [y=error, meta=label, col sep=comma,skip coords between index={0}{50}]{csv_synthetic4.csv};
\end{semilogxaxis}
\end{tikzpicture}
\hspace{6pt}
\begin{tikzpicture} 
\begin{semilogxaxis}[
	scale=1.0,
	xlabel={\# of Rules},
	xmin=1,
	xmax=500,
	ylabel style={align=center},
	ylabel={Test Error},
	ymin=0.05,
	ytick={0.05, 0.1, 0.15, 0.2},
	yticklabel style={
		/pgf/number format/fixed,
		/pgf/number format/precision=2,
		/pgf/number format/fixed zerofill
	},
	title style={yshift=-1.5ex,},
	title={Spambase},
	width=0.32\textwidth,
	height=4.5cm,
]
\addplot[
	scatter,only marks,scatter src=explicit symbolic,
	scatter/classes={
		defrag={mark=*,draw=black,fill=blue},
		BATree={mark=pentagon*,draw=black,fill=red},
		inTreesFull={mark=triangle*,draw=black,fill=green},
		NH={mark=square*,draw=black,fill=cyan},
		DTree2={mark=diamond*,draw=black,fill=magenta}
		}
	]
	table [x=rule, y=error, meta=label, col sep=comma,skip coords between index={50}{52}]{csv_spambase.csv};
	\addplot[mark=none, color=black, dashed] table [y=error, meta=label, col sep=comma,skip coords between index={0}{50}]{csv_spambase.csv};
\end{semilogxaxis}
\end{tikzpicture}

\begin{tikzpicture} 
\begin{semilogxaxis}[
	scale=1.0,
	xlabel={\# of Rules},
	xmin=1,
	xmax=500,
	ylabel style={align=center},
	ylabel={Test Error},
	ymin=0.08,
	ytick={0.08, 0.1, 0.12, 0.14, 0.16},
	yticklabel style={
		/pgf/number format/fixed,
		/pgf/number format/precision=2,
		/pgf/number format/fixed zerofill
	},
	title style={yshift=-1.5ex,},
	title={MiniBooNE},
	width=0.32\textwidth,
	height=4.5cm,
]
\addplot[
	scatter,only marks,scatter src=explicit symbolic,
	scatter/classes={
		defrag={mark=*,draw=black,fill=blue},
		BATree={mark=pentagon*,draw=black,fill=red},
		inTreesFull={mark=triangle*,draw=black,fill=green},
		NH={mark=square*,draw=black,fill=cyan},
		DTree2={mark=diamond*,draw=black,fill=magenta}
		}
	]
	table [x=rule, y=error, meta=label, col sep=comma,skip coords between index={50}{52}]{csv_miniboone.csv};
	\addplot[mark=none, color=black, dashed] table [y=error, meta=label, col sep=comma,skip coords between index={0}{50}]{csv_miniboone.csv};
\end{semilogxaxis}
\end{tikzpicture}
\hspace{6pt}
\begin{tikzpicture} 
\begin{semilogxaxis}[
	scale=1.0,
	xlabel={\# of Rules},
	xmin=1,
	xmax=500,
	ylabel style={align=center},
	ylabel={Test Error},
	ymin=0.3,
	ytick={0.3, 0.32, 0.34, 0.36, 0.38},
	yticklabel style={
		/pgf/number format/fixed,
		/pgf/number format/precision=2,
		/pgf/number format/fixed zerofill
	},
	title style={yshift=-1.5ex,},
	title={Higgs},
	width=0.32\textwidth,
	height=4.5cm,
]
\addplot[
	scatter,only marks,scatter src=explicit symbolic,
	scatter/classes={
		defrag={mark=*,draw=black,fill=blue},
		BATree={mark=pentagon*,draw=black,fill=red},
		inTreesFull={mark=triangle*,draw=black,fill=green},
		NH={mark=square*,draw=black,fill=cyan},
		DTree2={mark=diamond*,draw=black,fill=magenta}
		}
	]
	table [x=rule, y=error, meta=label, col sep=comma,skip coords between index={50}{52}]{csv_higgs.csv};
	\addplot[mark=none, color=black, dashed] table [y=error, meta=label, col sep=comma,skip coords between index={0}{50}]{csv_higgs.csv};
\end{semilogxaxis}
\end{tikzpicture}
\hspace{6pt}
\begin{tikzpicture} 
\begin{semilogxaxis}[
	scale=1.0,
	xlabel={\# of Rules},
	xmin=1,
	xmax=500,
	ylabel style={align=center},
	ylabel={Test Error},
	title style={yshift=-1.5ex,},
	title={Energy},
	width=0.32\textwidth,
	height=4.5cm,
]
\addplot[
	scatter,only marks,scatter src=explicit symbolic,
	scatter/classes={
		defrag={mark=*,draw=black,fill=blue},
		BATree={mark=pentagon*,draw=black,fill=red},
		inTreesFull={mark=triangle*,draw=black,fill=green},
		NH={mark=square*,draw=black,fill=cyan},
		DTree2={mark=diamond*,draw=black,fill=magenta}
		}
	]
	table [x=rule, y=error, meta=label, col sep=comma,skip coords between index={50}{52}]{csv_energy.csv};
	\addplot[mark=none, color=black, dashed] table [y=error, meta=label, col sep=comma,skip coords between index={0}{50}]{csv_energy.csv};
\end{semilogxaxis}
\end{tikzpicture}
\caption{Comparison of the simplification methods: \# of rules vs. test error. \textit{Ensemble} denotes the average error of the tree ensemble.}
\label{fig:result}
\end{figure*}

We compared the performance of the proposed method with the baseline methods with respect to the number of found rules and the test errors.
We conducted the experiment over ten random data realizations for each dataset.
\figurename~\ref{fig:result} shows the trade-off between the number of found rules $K$ and the test errors of each method on three datasets.

\paragraph{Number of Found Rules:}
\figurename~\ref{fig:result} shows that DTree2 tended to attain the smallest number of rules (i.e., four), and the proposed method was second (from three to ten).
The number of rules found by inTrees and Node Harvest tended to be around 30 to 100, while the number of rules found by BATrees sometimes exceeded 100.

\paragraph{Test Errors:}
\figurename~\ref{fig:result} also shows inTrees and BATrees tended to attain the smallest test errors while DTree2 appeared to perform the worst on most of the datasets.
The proposed method attained a good trade-off between the number of rules and the test errors: it tended to score smaller errors than DTree2 while using only a few rules, which is significantly smaller than the other baseline methods.

These results suggest that the proposed method is favorable for interpretation as it generates only a few rules with small test errors.
A smaller number of rules helps users to easily check the found rules.
The small test errors support that the found rules are reliable, i.e., the rules explain the original tree ensemble adequately.

\subsection{Example of Found Rules}

We show rule examples found in the experiment.

%

\begin{figure*}[t]
	\centering
	\subfigure[Original Data]{
		\includegraphics[width=0.28\textwidth]{synthetic4_true.pdf} 
		\label{fig:true_s1}}
	\subfigure[Learned Tree Ensemble]{
		\includegraphics[width=0.28\textwidth]{synthetic4_rf_tree05_seed00_787.pdf}
		\label{fig:rf_s1}}
	\subfigure[Proposed]{
		\includegraphics[width=0.28\textwidth]{synthetic4_defrag.pdf} 
		\label{fig:defrag_s1}}
	\subfigure[BATrees]{
		\includegraphics[width=0.28\textwidth]{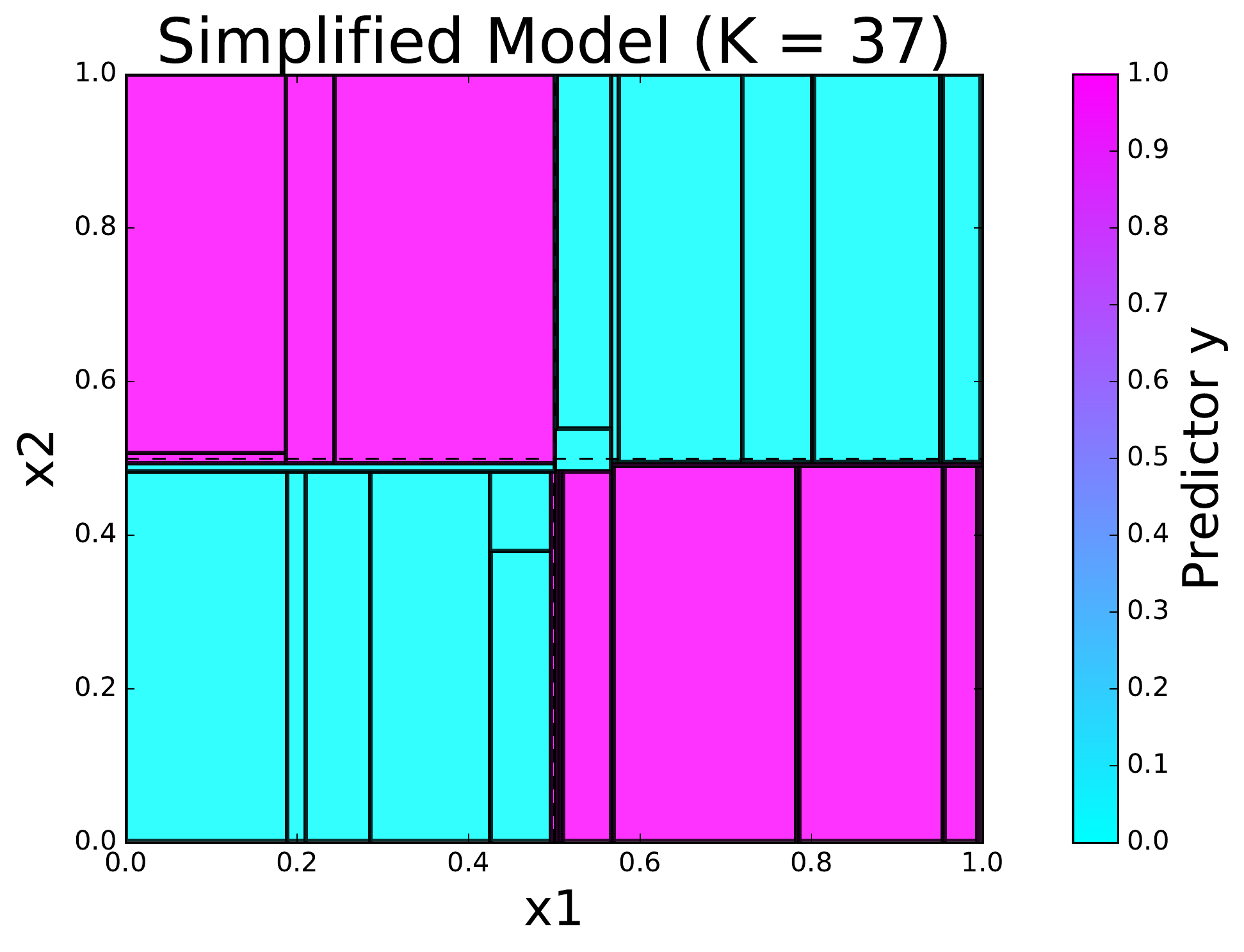} 
		\label{fig:ba_s1}}
	\subfigure[inTrees]{
		\includegraphics[width=0.28\textwidth]{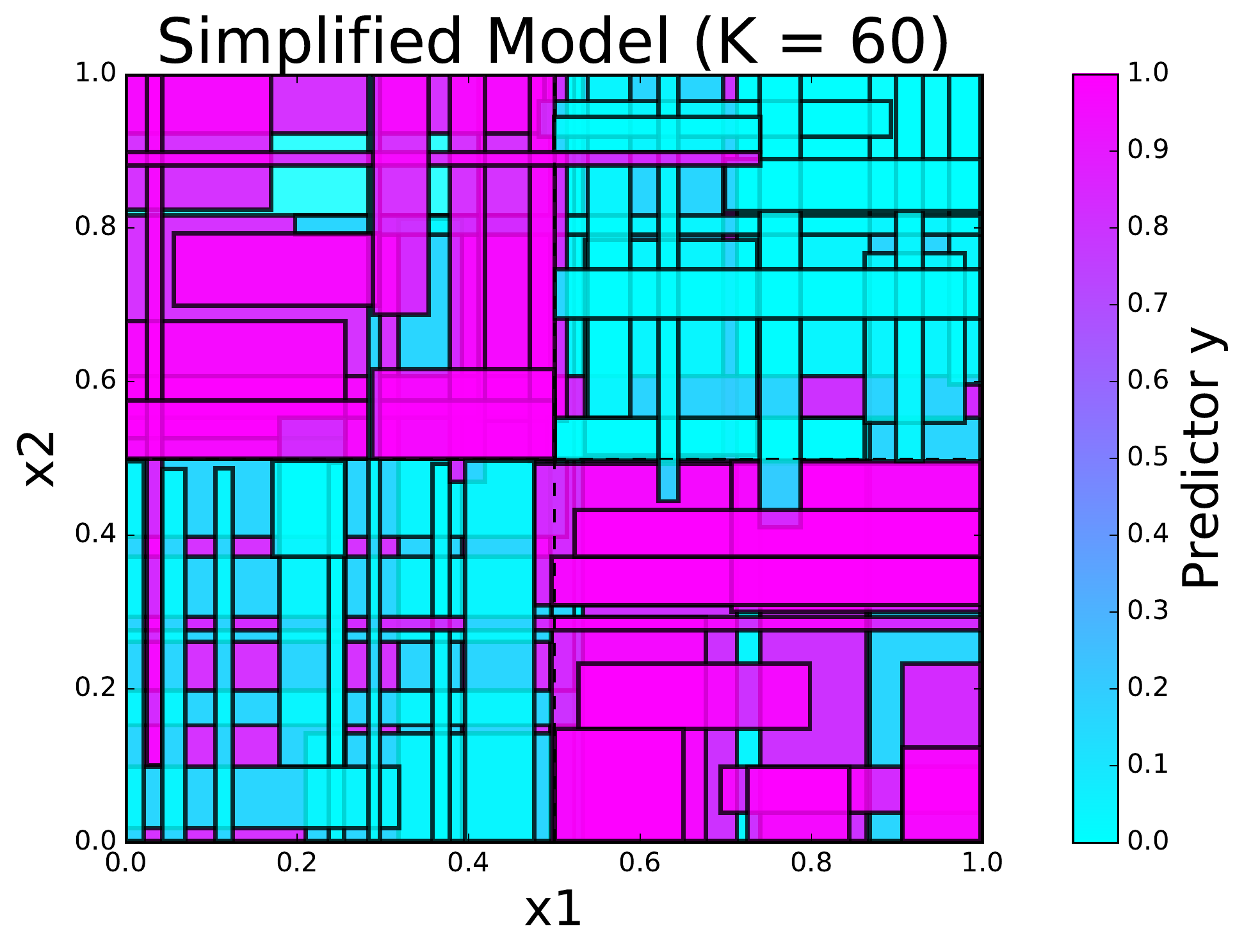}
		\label{fig:in_s1}}
	\subfigure[DTree2]{
		\includegraphics[width=0.28\textwidth]{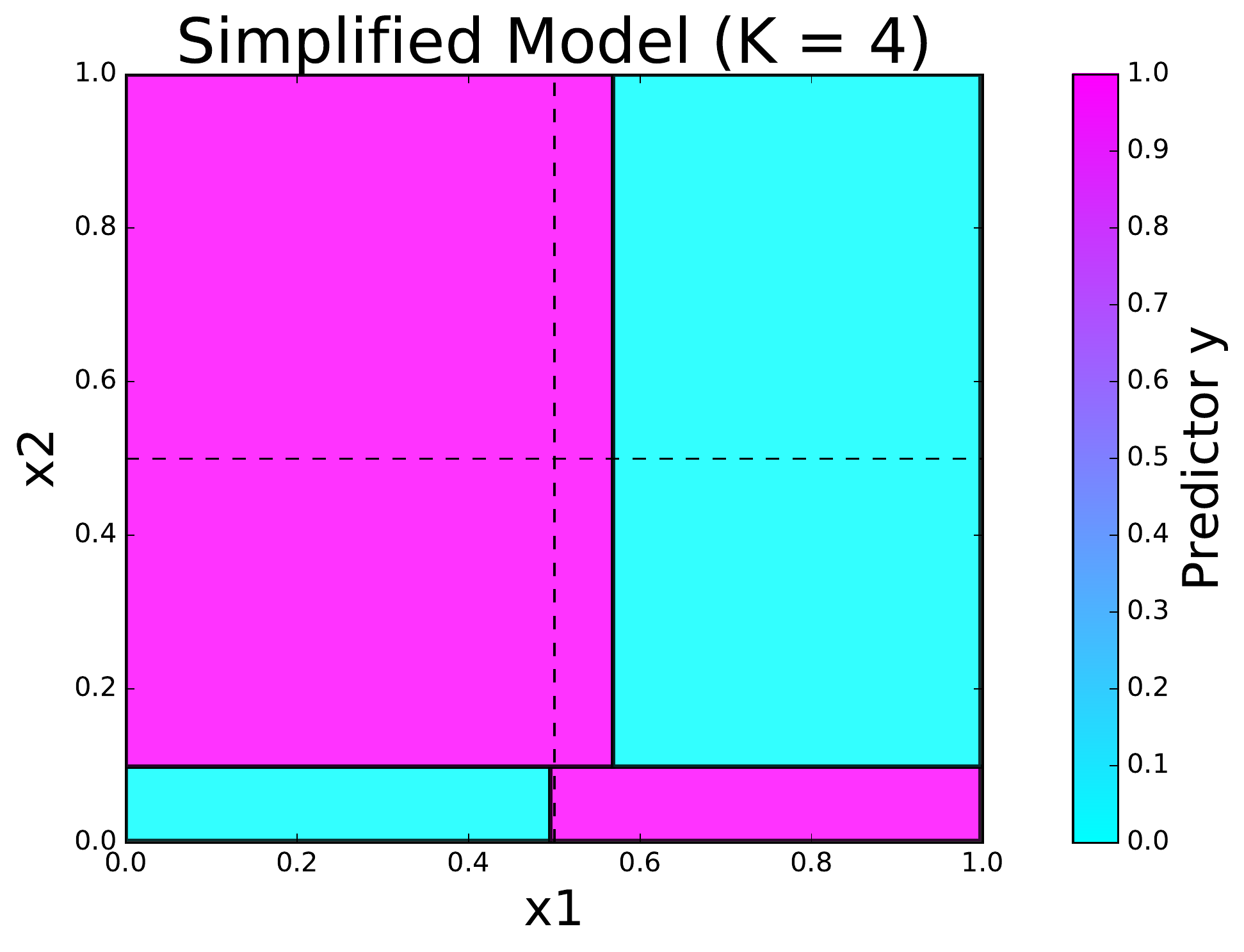} 
		\label{fig:d2_s1}}
	\caption{Synthetic1: Original data, leaned tree ensemble, and simplified rules.}
	\label{fig:synthetic1}
\end{figure*}

\begin{figure*}[!ht]
	\centering
	\subfigure[Original Data]{
		\includegraphics[width=0.28\textwidth]{synthetic3_true.pdf} 
		\label{fig:true_s2}}
	\subfigure[Learned Tree Ensemble]{
		\includegraphics[width=0.28\textwidth]{synthetic3_rf_tree05_seed00_1374.pdf}
		\label{fig:rf_s2}}
	\subfigure[Proposed]{
		\includegraphics[width=0.28\textwidth]{synthetic3_defrag.pdf} 
		\label{fig:defrag_s2}}
	\subfigure[BATrees]{
		\includegraphics[width=0.28\textwidth]{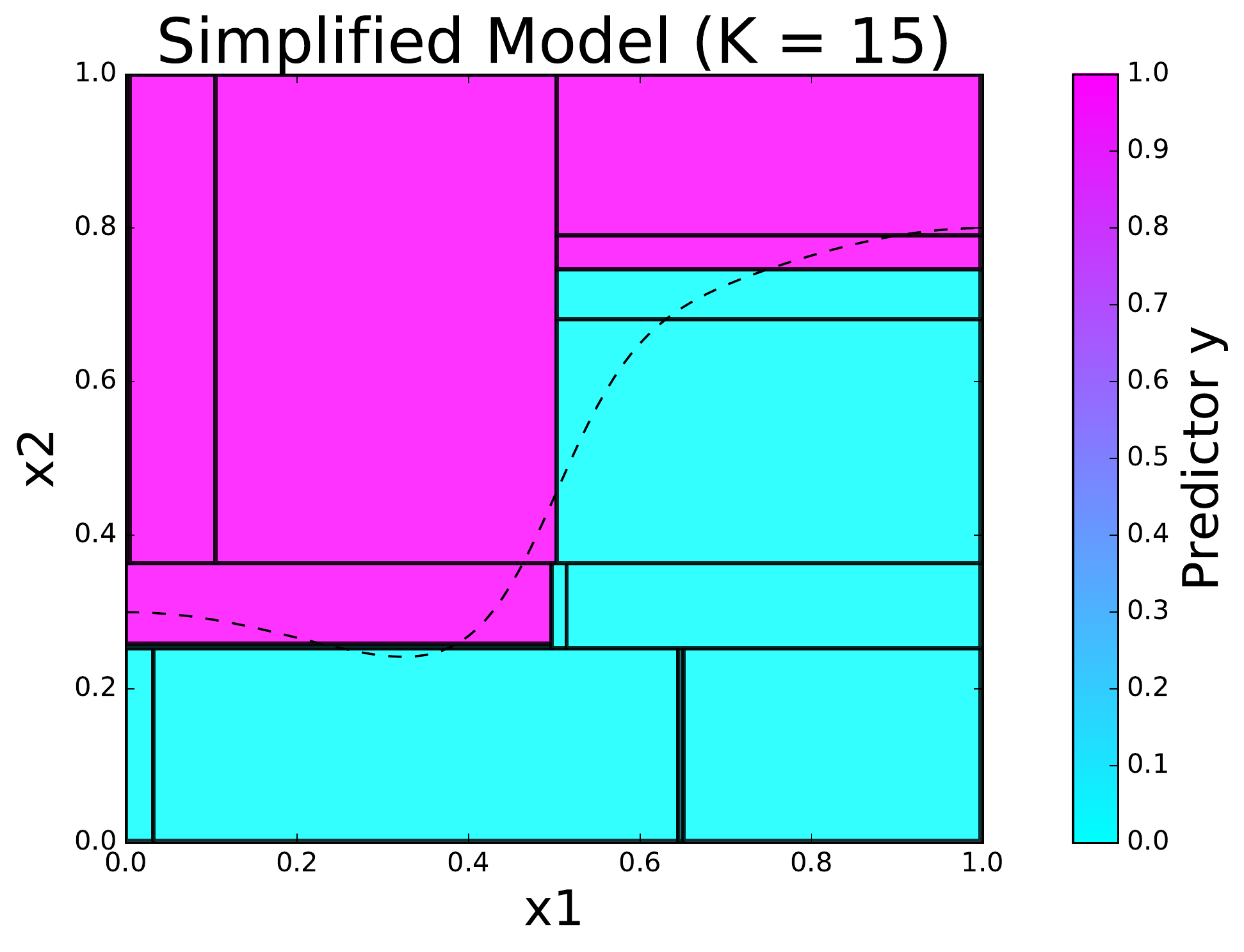} 
		\label{fig:ba_s2}}
	\subfigure[inTrees]{
		\includegraphics[width=0.28\textwidth]{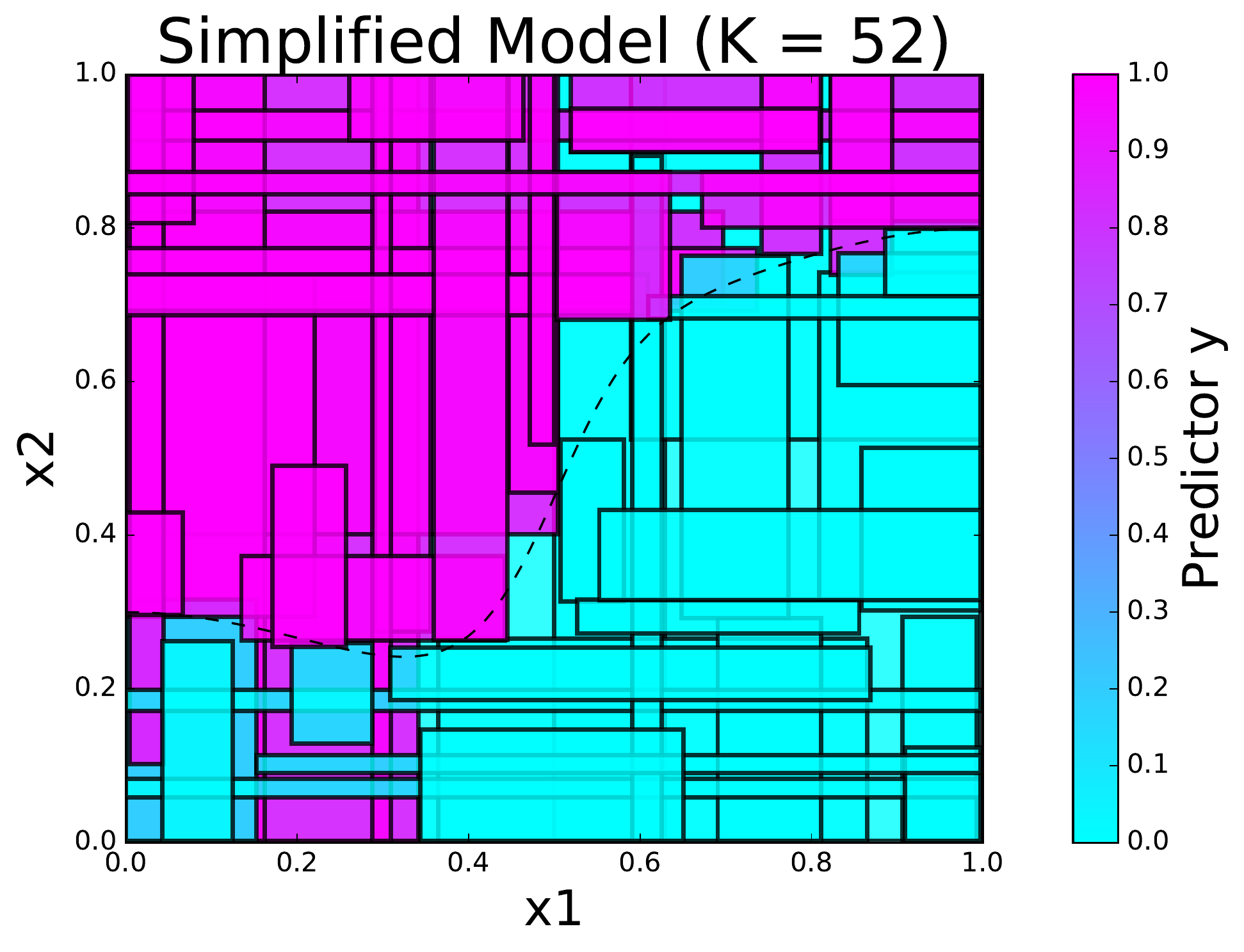}
		\label{fig:in_s2}}
	\subfigure[DTree2]{
		\includegraphics[width=0.28\textwidth]{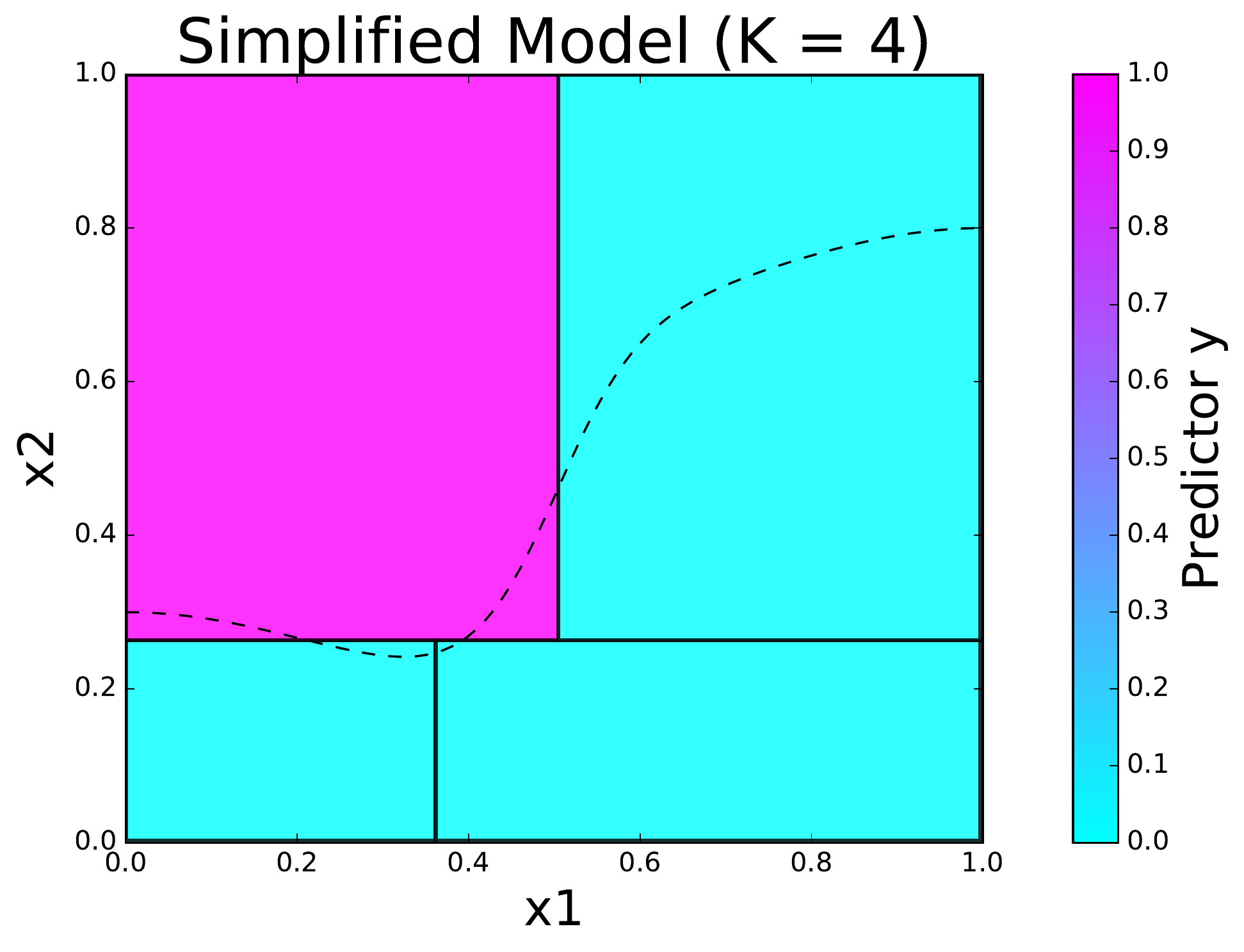} 
		\label{fig:d2_s2}}
	\caption{Synthetic2: Original data, leaned tree ensemble, and simplified rules.}
	\label{fig:synthetic2}
\end{figure*}

\begin{figure*}[!ht]
	\centering
	\includegraphics[width=0.8\textwidth]{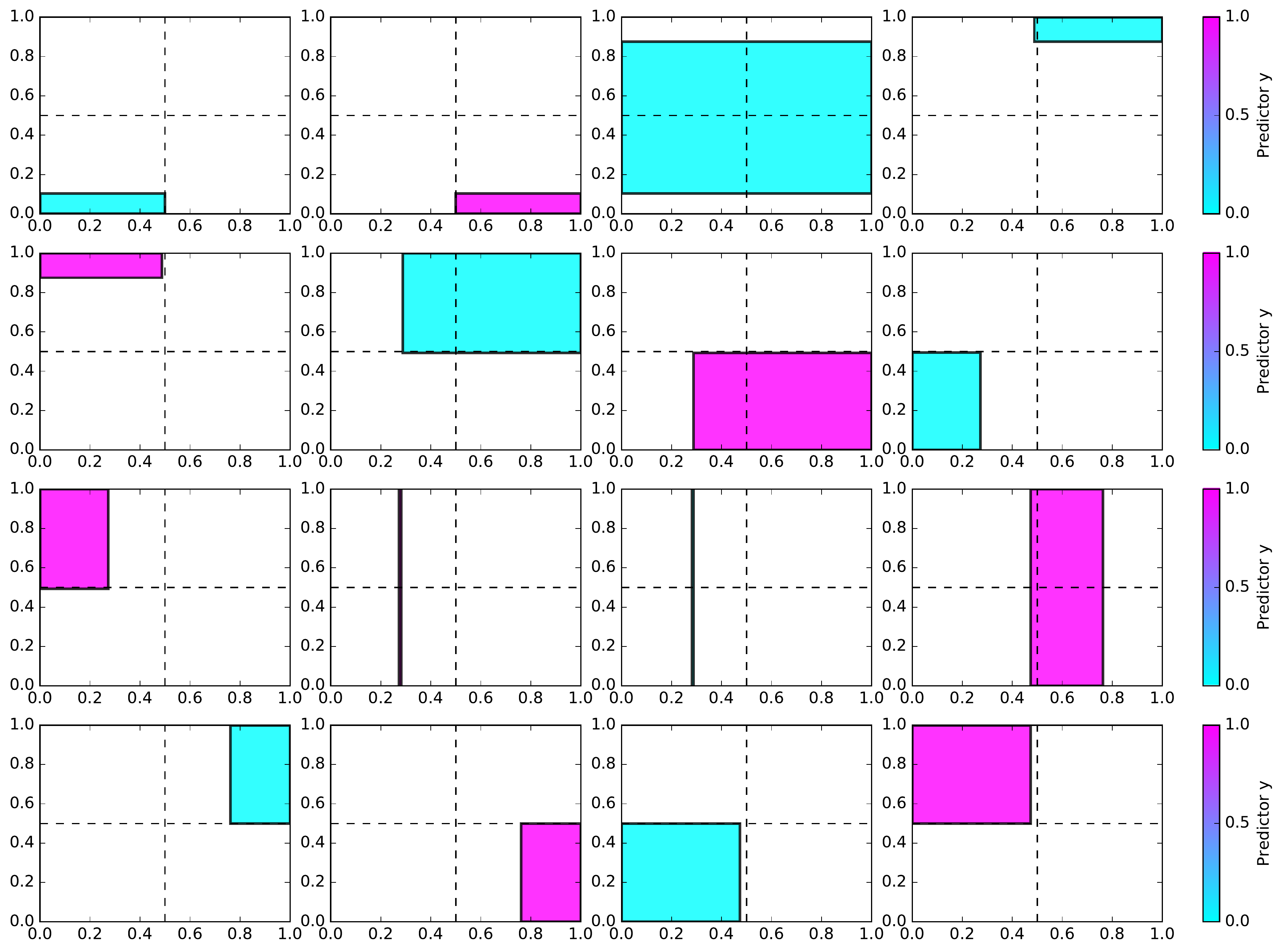} 
	\caption{Synthetic1: Found rules by Node Harvest.}
	\label{fig:synthetic1_nh}
\end{figure*}

\begin{figure*}[!ht]
	\centering
	\includegraphics[width=0.8\textwidth]{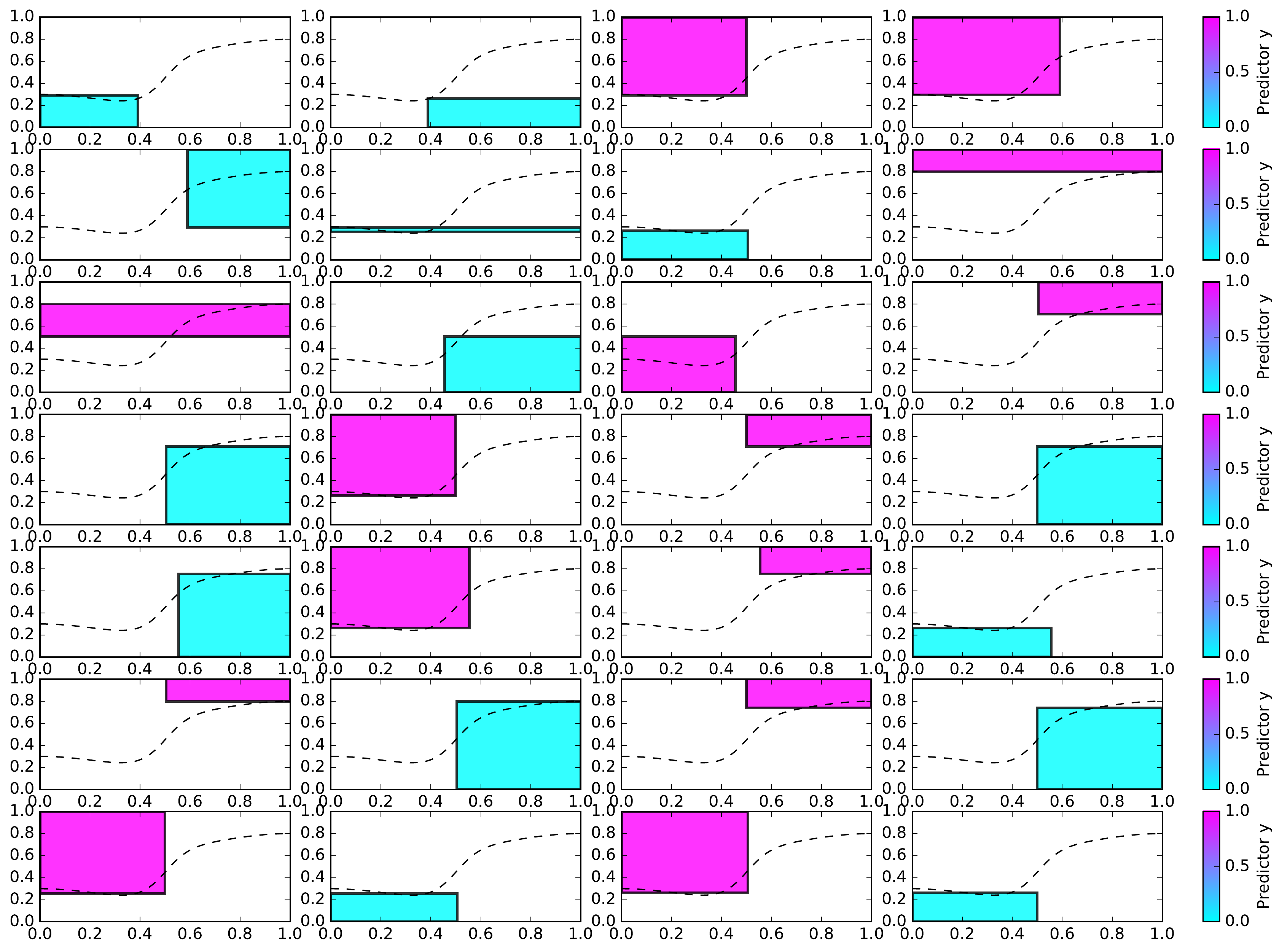} 
	\caption{Synthetic2: Found rules by Node Harvest.}
	\label{fig:synthetic2_nh}
\end{figure*}

\paragraph{Synthetic1 \& Synthetic2:} 
\figurename~\ref{fig:synthetic1} and \ref{fig:synthetic2} show the simplified rules of the learned tree ensemble on Synthetic1 and Synthetic2, respectively.
The results on Node Harvest can be found in \figurename~\ref{fig:synthetic1_nh} and \ref{fig:synthetic2_nh}.
The proposed method well simplified the boundary using only a few rules.
On the other hand, BATrees and inTrees required more rules.
It is important to note that inTrees and Node Harvest found rules that highly overlap each other.
The rule overlapping is not favorable for interpretation: if the prediction of the overlapping rules are distinct, we cannot decide which rule to trust.
Although there are some overlaps between the rules in the proposed method, this is not that critical as observed for inTrees and Node Harvest because the overlapped regions are limited.
\tablename~\ref{tab:app_coll} shows the average number of overlapped rules where the ideal value is one.
While the proposed method attained values close to one, inTrees and Node Harvest scored far larger values.

\begin{table*}[t]
	\centering
	\caption{Average Number of Rules Covering Each Test Point: BATrees is omitted because its value is always one.}
	\label{tab:app_coll}
	\begin{tabular}{ccccccc}
	& Synthetic1 & Synthetic2 & Spambase & MiniBooNE & Higgs & Energy \\
	\hline
   Proposed & $1.01 \pm 0.03$ & $1.05 \pm 0.06$ & $1.60 \pm 0.12$ & $2.51 \pm 0.09$ & $1.56 \pm 0.21$ & $0.95 \pm 0.08$ \\
   inTrees & $4.65 \pm 0.38$ & $3.73 \pm 0.41$ & $5.39 \pm 0.34$ & $6.25 \pm 0.28$ & $5.53 \pm 0.27$ & $3.30 \pm 0.19$ \\
	Node Harvest & $3.53 \pm 0.95$ & $8.78 \pm 2.02$ & $12.4 \pm 1.76$ & $17.9 \pm 2.24$ & $9.42 \pm 1.85$ & $3.61 \pm 0.19$
	\end{tabular}
\end{table*}

\paragraph{Energy Data:} 
Energy efficiency data is a simulation data sampled from 12 different building shapes.
The dataset comprises eight numeric features which are Relative Compactness, Surface Area, Wall Area, Roof Area, Overall Height, Orientation, Glazing Area, and Glazing Area Distribution.
The task is regression, which aims to predict the heating load of the building from these eight features.

In \tablename~\ref{tab:app_energy}, the four rules found by the proposed method are characterized by the two features Overall Height and Wall Area.
The four rules are expressed as a direct product of the two statements; (i) Overall Height $\in \{{\rm low}, {\rm high}\}$, and (ii) Wall Area $\in \{{\rm small}, {\rm large}\}$.
The resulting rules are intuitive such that the load is small when the building is small, while the load is large when the building is huge.
Hence, from these simplified rules, we can infer that the tree ensemble is learned in accordance with our intuition about the data.
In contrast to the simple rules found by the proposed method, the baseline methods found more rules: BATrees learned 66 rules, inTrees enumerated 23 rules, and Node Harvest found 10 rules, respectively.
\tablename~\ref{tab:app_energy} shows four example rules found by each method.

\begin{table*}[!ht]
	\small
	\caption{Examples of extracted rules using the tree ensemble simplification methods on Energy data.}
	\label{tab:app_energy}
	\centering
	\begin{tabular}{ccp{0.74\hsize}}
		$y$ &  Rule\\
		\hline
		\multirow{4}{*}{\rotatebox[origin=c]{90}{\shortstack{Proposed}}}
& $12.33$ &${\rm Overall Height} < 5.25$, ${\rm Wall Area} < 306.25$ \\
& $14.39$ &${\rm Overall Height} < 5.25$, ${\rm Wall Area} \geq 318.50$ \\
& $28.17$ &${\rm Overall Height} \geq 5.25$, ${\rm Wall Area} < 330.75$ \\
& $37.38$ &${\rm Overall Height} \geq 5.25$, ${\rm Wall Area} \geq 343.00$ \\
		\hline
		\multirow{8}{*}{\rotatebox[origin=c]{90}{\shortstack{BATrees}}}
& $17.18$ &${\rm Relative Compactness} < 0.84$, ${\rm Wall Area} < 330.75$, ${\rm Roof Area} < 183.75$, ${\rm Glazing Area} < 0.33$, ${\rm Glazing Area Distribution} < 0.50$ \\
& $24.50$ &${\rm Relative Compactness} < 0.84$, ${\rm Wall Area} < 330.75$, ${\rm Roof Area} < 183.75$, ${\rm Orientation} < 3.50$, ${\rm Glazing Area} < 0.33$, $0.50 \leq {\rm Glazing Area Distribution} < 3.50$ \\
& $24.20$ &${\rm Relative Compactness} < 0.84$, ${\rm Wall Area} < 330.75$, ${\rm Roof Area} < 183.75$, $3.50 \leq {\rm Orientation} < 4.50$, ${\rm Glazing Area} < 0.33$, $0.50 \leq {\rm Glazing Area Distribution} < 3.50$ \\
& $23.94$ &${\rm Relative Compactness} < 0.84$, ${\rm Wall Area} < 330.75$, ${\rm Roof Area} < 183.75$, ${\rm Orientation} \geq 4.50$, ${\rm Glazing Area} < 0.33$, $0.50 \leq {\rm Glazing Area Distribution} < 3.50$ \\
		\hline
		\multirow{4}{*}{\rotatebox[origin=c]{90}{\shortstack{inTrees}}}
& $12.58$ &${\rm Relative Compactness} \geq 0.65$, ${\rm Overall Height} < 5.25$, ${\rm Glazing Area} < 0.33$ \\
& $33.20$ &${\rm Relative Compactness} \geq 0.75$, ${\rm Surface Area} \geq 624.75$, ${\rm Glazing Area Distribution} \geq 0.50$ \\
& $33.20$ &${\rm Relative Compactness} \geq 0.84$, ${\rm Glazing Area} \geq 0.33$ \\
& $23.61$ &$673.75 \leq {\rm Surface Area} < 796.25$, ${\rm Wall Area} \geq 306.25$, ${\rm Glazing Area} \geq 0.17$ \\
		\hline
		\multirow{4}{*}{\rotatebox[origin=c]{90}{NH}}
& $14.53$ &${\rm Surface Area} \geq 674.00$, ${\rm Glazing Area} \geq 0.17$ \\
& $28.17$ &${\rm Relative Compactness} \geq 0.81$, ${\rm Surface Area} < 674.00$ \\
& $37.38$ &${\rm Relative Compactness} < 0.81$, ${\rm Surface Area} < 674.00$ \\
& $11.21$ &${\rm Surface Area} \geq 674.00$, ${\rm Glazing Area} < 0.17$ \\
		\hline
		\multirow{4}{*}{\rotatebox[origin=c]{90}{DTree2}}
& $11.21$ &${\rm Surface Area} \geq 673.75$, ${\rm Glazing Area} < 0.17$ \\
& $14.53$ &${\rm Surface Area} \geq 673.75$, ${\rm Glazing Area} \geq 0.17$ \\
& $28.17$ &${\rm Surface Area} < 624.75$ \\
& $37.38$ &$624.75 \leq {\rm Surface Area} < 673.75$
	\end{tabular}
\end{table*}

\section{Conclusion}
\label{sec:concl}

We proposed a simplification method for tree ensembles to enable users to interpret the model.
The difficulty of interpreting tree ensembles arises because the trees divide an input space into more than a thousand small regions.
We simplified the complex tree ensemble by optimizing the number of regions in the model.
We formalized this simplification as a model selection problem: Given a complex tree ensemble, we want to obtain the simplest representation that is essentially equivalent to the original one.
To solve this problem, we derived a Bayesian model selection algorithm that automatically determines the model complexity.
By using the proposed method, the complex ensemble is approximated with a simple model that is easy to interpret.
Our numerical experiments on several datasets showed that complicated tree ensembles were approximated adequately.

\clearpage

\bibliography{main}

\begin{thebibliography}{20}
\providecommand{\natexlab}[1]{#1}
\providecommand{\url}[1]{\texttt{#1}}
\expandafter\ifx\csname urlstyle\endcsname\relax
  \providecommand{\doi}[1]{doi: #1}\else
  \providecommand{\doi}{doi: \begingroup \urlstyle{rm}\Url}\fi

\bibitem[Breiman(2001)]{breiman2001random}
L.~Breiman.
\newblock Random forests.
\newblock \emph{Machine learning}, 45\penalty0 (1):\penalty0 5--32, 2001.

\bibitem[Friedman(2001)]{friedman2001greedy}
J.~H. Friedman.
\newblock Greedy function approximation: A gradient boosting machine.
\newblock \emph{Annals of statistics}, 29\penalty0 (5):\penalty0 1189--1232,
  2001.

\bibitem[Chen and Guestrin(2016)]{chen2016xgboost}
T.~Chen and C.~Guestrin.
\newblock Xgboost: A scalable tree boosting system.
\newblock \emph{Proceedings of the 22nd ACM SIGKDD International Conference on
  Knowledge Discovery and Data Mining}, pages 785--794, 2016.

\bibitem[Kaggle(2017)]{Kaggle2016}
Kaggle.
\newblock Your year on kaggle: Most memorable community stats from 2016, 2017.
\newblock URL
  \url{http://blog.kaggle.com/2017/01/05/your-year-on-kaggle-most-memorable-community-stats-from-2016/}.

\bibitem[Akaike(1974)]{akaike1974new}
H.~Akaike.
\newblock A new look at the statistical model identification.
\newblock \emph{IEEE transactions on automatic control}, 19\penalty0
  (6):\penalty0 716--723, 1974.

\bibitem[Schwarz(1978)]{schwarz1978estimating}
G.~Schwarz.
\newblock Estimating the dimension of a model.
\newblock \emph{The annals of statistics}, 6\penalty0 (2):\penalty0 461--464,
  1978.

\bibitem[Kass and Raftery(1995)]{kassr95}
R.~E. Kass and A.~E. Raftery.
\newblock Bayes factors.
\newblock \emph{Journal of the American Statistical Association}, 90:\penalty0
  773--795, 1995.

\bibitem[Fujimaki and Morinaga(2012)]{fujimaki2012factorized}
R.~Fujimaki and S.~Morinaga.
\newblock Factorized asymptotic bayesian inference for mixture modeling.
\newblock \emph{Proceedings of the 15th International Conference on Artificial
  Intelligence and Statistics}, pages 400--408, 2012.

\bibitem[Hayashi et~al.(2015)Hayashi, Maeda, and
  Fujimaki]{hayashi2015rebuilding}
K.~Hayashi, S.~Maeda, and R.~Fujimaki.
\newblock Rebuilding factorized information criterion: Asymptotically accurate
  marginal likelihood.
\newblock \emph{Proceedings of the 32nd International Conference on Machine
  Learning}, pages 1358--1366, 2015.

\bibitem[Kim et~al.(2016{\natexlab{a}})Kim, Malioutov, and
  Varshney]{kim2016proceedings}
B.~Kim, D.~M. Malioutov, and K.~R. Varshney.
\newblock Proceedings of the 2016 {ICML} workshop on human interpretability in
  machine learning.
\newblock \emph{arXiv preprint arXiv:1607.02531}, 2016{\natexlab{a}}.

\bibitem[Wilson et~al.(2016)Wilson, Kim, and Herlands]{wilson2016proceedings}
A.~G. Wilson, B.~Kim, and W.~Herlands.
\newblock Proceedings of {NIPS} 2016 workshop on interpretable machine learning
  for complex systems.
\newblock \emph{arXiv preprint arXiv:1611.09139}, 2016.

\bibitem[Ribeiro et~al.(2016)Ribeiro, Singh, and Guestrin]{ribeiro2016should}
M.~T. Ribeiro, S.~Singh, and C.~Guestrin.
\newblock Why {S}hould {I} {T}rust {Y}ou?: Explaining the predictions of any
  classifier.
\newblock \emph{Proceedings of the 22nd ACM SIGKDD International Conference on
  Knowledge Discovery and Data Mining}, pages 1135--1144, 2016.

\bibitem[Kim et~al.(2016{\natexlab{b}})Kim, Khanna, and
  Koyejo]{kim2016examples}
B.~Kim, R.~Khanna, and O.~Koyejo.
\newblock Examples are not enough, learn to criticize! {C}riticism for
  interpretability.
\newblock \emph{Advances In Neural Information Processing Systems}, pages
  2280--2288, 2016{\natexlab{b}}.

\bibitem[Kaufman et~al.(2012)Kaufman, Rosset, Perlich, and
  Stitelman]{kaufman2012leakage}
S.~Kaufman, S.~Rosset, C.~Perlich, and O.~Stitelman.
\newblock Leakage in data mining: Formulation, detection, and avoidance.
\newblock \emph{ACM Transactions on Knowledge Discovery from Data}, 6\penalty0
  (4):\penalty0 15, 2012.

\bibitem[Lloyd and Ghahramani(2015)]{lloyd2015statistical}
J.~R. Lloyd and Z.~Ghahramani.
\newblock Statistical model criticism using kernel two sample tests.
\newblock \emph{Advances in Neural Information Processing Systems}, pages
  829--837, 2015.

\bibitem[Kim et~al.(2015)Kim, Shah, and Doshi-Velez]{kim2015mind}
B.~Kim, J.~Shah, and F.~Doshi-Velez.
\newblock Mind the gap: {A} generative approach to interpretable feature
  selection and extraction.
\newblock \emph{Advances in Neural Information Processing Systems}, pages
  2260--2268, 2015.

\bibitem[Breiman and Shang(1996)]{breiman1996born}
L.~Breiman and N.~Shang.
\newblock Born again trees.
\newblock \emph{University of California, Berkeley, Berkeley, CA, Technical
  Report}, 1996.

\bibitem[Deng(2014)]{deng2014interpreting}
H.~Deng.
\newblock Interpreting tree ensembles with intrees.
\newblock \emph{arXiv preprint arXiv:1408.5456}, 2014.

\bibitem[Meinshausen(2010)]{meinshausen2010node}
N.~Meinshausen.
\newblock Node harvest.
\newblock \emph{The Annals of Applied Statistics}, 4\penalty0 (4):\penalty0
  2049--2072, 2010.

\bibitem[Lichman(2013)]{Lichman:2013}
M.~Lichman.
\newblock {UCI} machine learning repository, 2013.
\newblock URL \url{http://archive.ics.uci.edu/ml}.

\end{thebibliography}
\bibliographystyle{abbrvnat}

\clearpage

\section*{Appendix}
\appendix
\section{EM Algorithm}
\label{sec:em}

For fixed $K$, we can estimate the model parameter $\Pi$ using the maximum likelihood estimation:
\begin{align}
	\max_{\Pi} \log p(\mathcal{D} | \Pi, K) .
	\label{eq:mle}
\end{align}

The optimization problem (\ref{eq:mle}) is solved by the EM algorithm.
The lower bound of (\ref{eq:mle}) is derived as
\begin{align}
	\sum_{n=1}^N \sum_{k=1}^K \mathbb{E}_{q(U)}[u_k^{(n)}] \log p(y^{(n)} | k, \phi) p(\bm{s}^{(n)} | k, \eta) p(k | \alpha) + H(q(U)) ,
	\label{eq:em_lb}
\end{align}
where $q(U)$ is the distribution of $U$, and $H(q(U))$ is an entropy of $q(U)$.
The EM algorithm is then formulated as an alternating maximization with respect to $q$ (E-step) and the parameter $\Pi$ (M-step).

\paragraph{[E-Step]}
In E-Step, we fix the parameter $\Pi$ and maximize the lower bound (\ref{eq:em_lb}) with respect to the distribution $q(U)$, which yields
\begin{align}
	q(u_k^{(n)} = 1) \propto p(y^{(n)} | k, \phi) p(\bm{s}^{(n)} | k, \eta) p(k | \alpha) .
	\label{eq:em_estep}
\end{align}

\paragraph{[M-Step]}
In M-Step, we fix the value of $q(u_k^{(n)}) = \beta_k^{(n)}$, and maximize the lower bound (\ref{eq:em_lb}) with respect to the parameter $\Pi$.
We then have, for $\eta$ and $\alpha$, 
\begin{align*}
	\eta_{k \ell} = \frac{\sum_{n=1}^N \beta_k^{(n)} s_{\ell}^{(n)}}{\sum_{n=1}^N \beta_k^{(n)}}, \qquad \alpha_k = \frac{1}{N} \sum_{n=1}^N \beta_k^{(n)} .
\end{align*}
The parameter $\phi$ is also updated as
\begin{align*}
	& \mu_k = \frac{\sum_{n=1}^N \beta_k^{(n)} z^{(n)}}{\sum_{n=1}^N \beta_k^{(n)}}, \qquad \lambda_k = \frac{\sum_{n=1}^N \beta_k^{(n)}}{\sum_{n=1}^N \beta_k^{(n)} (z^{(n)} - \mu_k)^2}, \\
	& \gamma_{kc} = \frac{\sum_{n=1}^N \beta_k^{(n)} z_c^{(n)}}{\sum_{n=1}^N \beta_k^{(n)}} .
\end{align*}

\section{FAB Lower Bound}
\label{sec:lb}

Here, we derive the lower bound of the marginal log-likelihood.
\begin{theorem}
	\label{th:lb}
	The marginal log-likelihood $\log p(\mathcal{D} | K)$ is lower bounded by (\ref{eq:lb}) except the $O(1)$ term.
\end{theorem}
The proof of this theorem follows the next three lemmas.

\begin{lemma}
	\label{lem:lb1}
	Let $U = \{ \bm{u}^{(n)} \}_{n=1}^{N}$ and the complete data likelihood be $p(\mathcal{D}, U | \Pi, K) = \prod_{n} p(y^{(n)}, \bm{s}^{(n)}, \bm{u}^{(n)} | \Pi, K)$.
	The marginal log-likelihood $\log p(\mathcal{D} | K)$ is lower bounded by
  \begin{align}
	\mathbb{E}_{q^(U)} \left[ \log p(\mathcal{D}, U | \hat{\Pi}, K) - \frac{1}{2} \log \det F_{\hat{\Pi}} \right] + H(q(U)) - \frac{{\rm dim} \hat{\Pi}}{2} \log N + O(1) ,
	\label{eq:fic2}
  \end{align}
where $q(U)$ is the distribution of $U$, $\hat{\Pi} = \argmaxl_{\Pi} \log p(\mathcal{D}, U | \Pi, K)$, $F_{\hat{\Pi}}$ is the Hessian of $- \log p(\mathcal{D}, U | \Pi, K) / N$ at $\Pi = \hat{\Pi}$, and $O(1)$ is the tern independent of $N$.
\end{lemma}

\noindent
(proof) Let $p(\mathcal{D}, U | K) = \int p(\mathcal{D}, U | \Pi, K) p(\Pi) d\Pi$.
From the definition of $\log p(\mathcal{D} | K)$, the next equation holds:
\begin{align}
	\log p(\mathcal{D} | K) =& \mathbb{E}_{q(U)} \left[ \log p(\mathcal{D}, U | K) \right] + H(q(U)) + {\rm KL}[q(U) || q^*(U)] ,
	\label{eq:lb1}
\end{align}
where ${\rm KL}[q(U) || q^*(U)]$ is a KL-divergence defined as
\begin{align}
	{\rm KL}[q(U) || q^*(U)] = \mathbb{E}_{q(U)} \left[ \log \frac{q(U)}{q^*(U)} \right] .
\end{align}
We note that the equation (\ref{eq:lb1}) can be easily verified from the next relationship
\begin{align}
	& {\rm KL}[q(U) || q^*(U)] \nonumber \\
	& = \mathbb{E}_{q(U)}[\log q(U)] - \mathbb{E}_{q(U)}[\log \underbrace{q^*(U)}_{= p(U | \mathcal{D}, K) = \frac{p(\mathcal{D}, U | K)}{p(\mathcal{D} | K)}}] \nonumber \\
	& = - H(q(U)) - \mathbb{E}_{q(U)}[\log p(\mathcal{D}, U | K)] + \log p(\mathcal{D} | K) .
\end{align}
Because the KL-divergence is non-negative, we have the lower bound of the marginal log-likelihood as
\begin{align}
	\log p(\mathcal{D} | K) \geq \mathbb{E}_{q(U)} \left[ \log p(\mathcal{D}, U | K) \right] + H(q(U)) .
	\label{eq:lb2}
\end{align}
We now apply Laplace's method to $\log p(\mathcal{D}, U | K)$, and derive the next equation~\cite{hayashi2015rebuilding}:
\begin{align}
	\begin{split}
	\log p(\mathcal{D}, U | K) = \log p(\mathcal{D}, U | \hat{\Pi}, K) - \frac{1}{2} \log \det F_{\hat{\Pi}} - \frac{{\rm dim} \hat{\Pi}}{2} \log N + O(1) .
	\end{split}
\end{align}
By substituting this result to (\ref{eq:lb2}), and we derive the lower bound (\ref{eq:fic2}).
\hfill $\Box$

\begin{lemma}
	\label{lem:lb2}
	The next inequality holds for any $\Pi$:
	\begin{align}
    		\mathbb{E}_{q(U)} \left[\log p(\mathcal{D}, U | \hat{\Pi}, K) \right] & \geq \mathbb{E}_{q(U)} \left[\log p(\mathcal{D}, U | \Pi, K) \right] \nonumber \\
		& = \sum_{n=1}^N \sum_{k=1}^K \mathbb{E}_{q(U)}[u_k^{(n)}] \log p(y^{(n)} | k, \phi) p(\bm{s}^{(n)} | k, \eta) p(k | \alpha) ,
		\label{eq:ineq1}
	\end{align}
	where $\hat{\Pi} = \argmaxl_{\Pi} \log p(\mathcal{D}, U | \Pi, K)$.
\end{lemma}

\noindent
(proof) It directly follows from the definition of $\hat{\Pi}$.
\hfill $\Box$

\begin{lemma}
	\label{lem:lb3}
	The next inequality holds:
	\begin{align}
		\begin{split}
		- \mathbb{E}_{q^(U)} \left[\frac{1}{2} \log \det F_{\hat{\Pi}} \right] - \frac{{\rm dim} \hat{\Pi}}{2} \log N \geq - \omega \sum_{k=1}^K \log \left( \sum_{n=1}^N \mathbb{E}_{q^(U)}[u_k^{(n)}] + 1 \right) + O(1) ,
		\end{split} \label{eq:ineq2}
	\end{align}
	where $\omega = ({\rm dim} \phi / K + L + 1) / 2$.
\end{lemma}

\noindent
(proof)  By expanding $\log \det F_{\Pi}$, we obtain
\begin{align}
	& \log \det F_{\Pi} \nonumber \\
	=& \sum_{k=1}^K \underbrace{\log \det \left( - \frac{\partial^2}{\partial^2 \phi} \frac{1}{N} \sum_{n=1}^N u_k^{(n)} \log p(y^{(n)} | k, \phi) \right)}_{(A)} \nonumber \\
	&+ \sum_{k=1}^K \underbrace{\log \det \left( - \frac{\partial^2}{\partial^2 \eta} \frac{1}{N} \sum_{n=1}^N u_k^{(n)} \log p(\bm{s}^{(n)} | k, \eta) \right)}_{(B)} \nonumber \\
	&+ \sum_{k=1}^K \underbrace{\log \det \left( - \frac{\partial^2}{\partial^2 \alpha} \frac{1}{N} \sum_{n=1}^N u_k^{(n)} \log p(k | \alpha) \right)}_{(C)} .
	\label{eq:fab_lb3}
\end{align}
Here, we have
\begin{align}
	(A) =& \frac{\dim \phi}{K} \log \left( \frac{1}{N} \sum_{n=1}^N u_k^{(n)} \right) + \log \det H_k \nonumber \\
    =& \underbrace{\frac{\dim \phi}{K} \log \left( \sum_{n=1}^N u_k^{(n)} \right) - \frac{\dim \phi}{K} \log N}_{O(\log N)} + \underbrace{\log \det H_k}_{O(1)} , 
	\label{eq:fab_lb3A}
\end{align}
\begin{align}
	(B) =& \sum_{\ell=1}^L \log \left( \frac{1}{N} \sum_{n=1}^N u_k^{(n)} \left( \frac{1}{\eta_{k\ell}^2} + \frac{1}{(1 - \eta_{k\ell})^2} \right) \right) \nonumber \\
	=& \underbrace{L \log \left( \sum_{n=1}^N u_k^{(n)} \right) - L \log N}_{O(\log N)} + \underbrace{\sum_{\ell=1}^L \log \left( \frac{\sum_{n=1}^N u_k^{(n)} \left( \frac{1}{\eta_{k\ell}^2} \frac{1}{(1 - \eta_{k\ell})^2} \right)}{\sum_{n=1}^N u_k^{(n)}} \right)}_{O(1)} ,
	\label{eq:fab_lb3B}
\end{align}
and
\begin{align}
	(C) =& \log \left( \frac{1}{N} \sum_{n=1}^N u_k^{(n)} \frac{1}{\alpha_k^2} \right) \nonumber \\
    =& \underbrace{\log \left( \sum_{n=1}^N u_k^{(n)} \right) - \log N}_{O(\log N)} - \underbrace{\log \alpha_k^2}_{O(1)} ,
	\label{eq:fab_lb3C}
\end{align}
where $O(1)$ denotes terms independent of $N$.
The matrix $H_k$ is given as, for the regression case, 
\begin{align*}
	H_k = \matrix{\lambda_k & \frac{\sum_{n=1}^N u_k^{(n)} (\mu_k - y^{(n)})}{\sum_{n=1}^N u_k^{(n)}} \\ \frac{\sum_{n=1}^N u_k^{(n)} (\mu_k - y^{(n)})}{\sum_{n=1}^N u_k^{(n)}} & \frac{1}{2\lambda_k^2}}
\end{align*}
and for the classification case,
\begin{align*}
	H_k = {\rm diag} \left( \frac{1}{\gamma_{k1}^2}, \frac{1}{\gamma_{k2}^2}, \ldots, \frac{1}{\gamma_{kC}^2} \right).
\end{align*}
By using these results, we can express $\log \det F_{\Pi}$ as
\begin{align}
	\log \det F_{\Pi} =& 2 \omega \sum_{k=1}^K \log \left( \sum_{n=1}^N u_k^{(n)} \right) - {\rm dim}\Pi \log N + O(1) .
\end{align}
We note that the only $O(1)$ term depends on $\Pi$ and the first two terms are independent of the value of $\Pi$.
Hence, the next equation holds for arbitrary $\Pi$:
\begin{align}
	- \frac{1}{2} \log \det F_{\hat{\Pi}} - \frac{{\rm dim}\hat{\Pi}}{2} \log N &= -\frac{1}{2} \log \det F_{\Pi} - \frac{{\rm dim} \Pi}{2} \log N + O(1) \nonumber \\
	& = - \omega \sum_{k=1}^K \log \left( \sum_{n=1}^N u_k^{(n)} \right) + O(1) .
\end{align}
Hence, the lower bound of $- \mathbb{E}_{q^*(U)} \left[\frac{1}{2} \log \det F_{\hat{\Pi}} \right] - \frac{{\rm dim} \hat{\Pi}}{2} \log N$ can be derived as
\begin{align}
	- \mathbb{E}_{q(U)} \left[\frac{1}{2} \log \det F_{\hat{\Pi}} \right] - \frac{{\rm dim} \hat{\Pi}}{2} \log N &= - \omega \sum_{k=1}^K \mathbb{E}_{q(U)} \left[ \log \left( \sum_{n=1}^N u_k^{(n)} \right) \right] + O(1) \nonumber \\
	& \geq - \omega \sum_{k=1}^K \log \left( \sum_{n=1}^N \mathbb{E}_{q(U)}[u_k^{(n)}] + 1 \right) + O(1) ,
\end{align}
where we used Jensen' inequality.
\hfill $\Box$

By using these lemmas, we now prove our main claim.

\noindent
(proof of Theorem~\ref{th:lb})
By substituting (\ref{eq:ineq1}) and (\ref{eq:ineq2}) into (\ref{eq:fic2}) and removing the $O(1)$ term, the claim follows.
\hfill $\Box$

\section{FAB Inference Algorithm Derivation}
\label{sec:derivation}

\paragraph{[E-Step]}
In E-Step, we update the distribution $q(U)$ so that the lower bound (\ref{eq:lb}) to be maximized.
Let $\beta_k^{(n)} = \mathbb{E}_{q(U)}[u_k^{(n)}] = q(u_k^{(n)})$.
The maximization problem can then be expressed as
\begin{align}
	\max_{\beta} \sum_{n=1}^N \sum_{k=1}^K \beta_k^{(n)} \log f_k^{(n)} - \omega \sum_{k=1}^K \log \left( \sum_{n=1}^N \beta_k^{(n)} + 1 \right) - \sum_{n=1}^N \sum_{k=1}^K \beta_k^{(n)} \log \beta_k^{(n)} , \;\; {\rm s.t.} \; \sum_{k=1}^K \beta_k^{(n)} = 1 ,
	\label{eq:app_estep}
\end{align}
where $f_k^{(n)} = p(y^{(n)} | k, \phi) p(\bm{s}^{(n)} | k, \eta) p(k | \alpha)$.
We note that the problem (\ref{eq:app_estep}) is smooth concave maximization, and a unique global optimum exists.
Such an optimum can be found by iterative maximization of the lower bound of (\ref{eq:app_estep}).
Recall that $\log \left( \sum_{n=1}^N \beta_k^{(n)} + 1 \right) \leq \log \left( \sum_{n=1}^N \psi_k^{(n)} + 1 \right) + \sum_{n=1}^N \frac{\beta_k^{(n)} - \psi_k^{(n)}}{\sum_{n'=1}^N \psi_k^{(n')} + 1}$ holds for any $\psi_k^{(n)}$ from the concavity.
Once the value of $\psi_k^{(n)}$ is fixed, we can maximize the lower bound of (\ref{eq:estep}) by solving
\begin{align}
	\max_{\beta} & \sum_{n=1}^N \sum_{k=1}^K \beta_k^{(n)} \log f_k^{(n)} - \omega \sum_{k=1}^K \sum_{n=1}^N \frac{\beta_k^{(n)}}{\sum_{n'=1}^N \psi_k^{(n')} + 1} - \sum_{n=1}^N \sum_{k=1}^K \beta_k^{(n)} \log \beta_k^{(n)} , \;\; {\rm s.t.} \; \sum_{k=1}^K \beta_k^{(n)} = 1 ,
	\label{eq:app_estep_lb}
\end{align}
which results in
\begin{align}
	\beta_k^{(n)} \propto f_k^{(n)} \exp\left( - \frac{\omega}{\sum_{n=1}^N \psi_k^{(n)} + 1} \right) ,
	\label{eq:app_estep_update}
\end{align}
Using this result, we can solve the original maximization problem (\ref{eq:app_estep}) by iteratively setting $\psi_k^{(n)} \leftarrow \beta_k^{(n)}$ and updating $\beta$ by (\ref{eq:app_estep_update}).
Because the lower bound (\ref{eq:app_estep_lb}) increases in every iteration, the iteration procedure converges to the global optimum.

\paragraph{[M-Step]}
In M-Step, we update the parameter $\Pi$ so that the lower bound (\ref{eq:lb}) to be maximized.
Let $\beta_k^{(n)} = \mathbb{E}_{q(U)}[u_k^{(n)}] = q(u_k^{(n)})$.
The maximization problem (\ref{eq:lb}) can then be decomposed into subproblems:
\begin{align}
	& \max_\phi \sum_{n=1}^N \beta_k^{(n)} \log p(y^{(n)} | k, \phi) , \label{eq:app_phi} \\
   & \max_\eta \sum_{n=1}^N \beta_k^{(n)} \log \eta_{k \ell}^{s_\ell^{(n)}} (1 - \eta_{k \ell})^{1 - s_\ell^{(n)}} , \label{eq:app_eta} \\
	& \max_\alpha \sum_{k=1}^K \left( \sum_{n=1}^N \beta_k^{(n)} \right) \log \alpha_k, \; {\rm s.t.} \; \sum_{k=1}^K \alpha_k = 1 , \label{eq:app_alpha} .
\end{align}
These problems can be solved analytically.
The solution to the problem (\ref{eq:app_phi}) are derived as
\begin{align}
	& \text{(regression):}
	\begin{cases}
    	& \mu_k = \frac{\sum_{n=1}^N \beta_k^{(n)} y^{(n)}}{\sum_{n=1}^N \beta_k^{(n)}}, \\
		& \lambda_k = \frac{\sum_{n=1}^N \beta_k^{(n)}}{\sum_{n=1}^N \beta_k^{(n)} (y^{(n)} - \mu_k)^2} ,
   \end{cases} \\
   & \text{(classification):} \;\;\; \gamma_{kc} = \frac{\sum_{n=1}^N \beta_k^{(n)} y_c^{(n)}}{\sum_{n=1}^N \beta_k^{(n)}}.
\end{align}
The solutions to the problem (\ref{eq:app_eta}) and (\ref{eq:app_alpha}) are derived as
\begin{align}
	\eta_{k \ell} = \frac{\sum_{n=1}^N \beta_k^{(n)} s_{\ell}^{(n)}}{\sum_{n=1}^N \beta_k^{(n)}} , \qquad \alpha_k = \frac{1}{N} \sum_{n=1}^N \beta_k^{(n)} .
\end{align}

\end{document}